\documentclass[twocolumn]{article}

\usepackage{amsmath}
\usepackage{amssymb}
\usepackage{bbm} % blackboard fonts, including for digits
\usepackage{bm} % Access bold symbols in maths mode
\usepackage{booktabs} % for better tables
\usepackage{dsfont} % for nicer blackboard bold \mathds{1}
\usepackage{graphicx}
\usepackage{mathtools}
\usepackage{microtype} % small typesetting improvements
\usepackage{multirow,array} % for strategic form games
\usepackage{soul} % for underlining with \ul
\usepackage{tabularx}
\usepackage{url}
\usepackage{xcolor}

\colorlet{mylinkcolor}{red}
\colorlet{mycitecolor}{green}
\colorlet{myurlcolor}{magenta}

\usepackage[
    colorlinks  = true,
    linkcolor   = mylinkcolor!80!black,
    citecolor   = mycitecolor!60!black,
    urlcolor    = myurlcolor!70!black,
]{hyperref}
\usepackage[margin=1in]{geometry}
\usepackage[small]{caption}
\usepackage[toc,page]{appendix}
\usepackage[utf8]{inputenc}    % utf8 support      

\usepackage{lmodern}

\usepackage[capitalize, nameinlink]{cleveref}

\usepackage{enumitem} % for better lists

\usepackage{tikz}
\usepackage{forest}
\usetikzlibrary{arrows, arrows.meta} % for prettier arrows
\usetikzlibrary{calc, backgrounds, positioning}

\crefname{section}{Sec.}{Secs.}
\crefname{appendix}{App.}{Apps.}
\crefname{figure}{Fig.}{Figs.}

\usepackage{ntheorem} % For theorem and definition environments
\theorembodyfont{\upshape}
\crefname{theorem}{Thm.}{Thms.}

\crefname{corollary}{Cor.}{Cors.}

\crefname{lemma}{Lem.}{Lems.}
\theoremstyle{definition}
\newtheorem{definition}{Definition}%[section]
\crefname{definition}{Def.}{Defs.}
\theoremstyle{remark}
\newtheorem{example}{Example}%[section]
\crefname{example}{Ex.}{Exs.}

\usepackage{apptools} % for appendix tools
\AtAppendix{\renewtheorem{definition}{Definition}[section]}

\usepackage[sorting=none,style=phys]{biblatex}
\AtEveryBibitem{%
\ifthenelse{
    \ifentrytype{article}\OR
    \ifentrytype{report}\OR
    \ifentrytype{collection}\OR
    \ifentrytype{incollection}\OR
    \ifentrytype{inproceedings}
    }{
    \clearfield{url}%
    \clearfield{urlyear}%
    \clearfield{review}%
    \clearfield{series}%%
    \clearfield{isbn}%%
    \clearfield{issn}%%
    \clearfield{doi}%%
    \clearfield{note}%%
    \clearfield{location}%%
    \clearfield{place}%%
}{}
\ifentrytype{book}{
    \clearfield{url}%
    \clearfield{urlyear}%
    \clearfield{review}%
    \clearfield{series}%%
    \clearfield{isbn}%%
    \clearfield{issn}%%
    \clearfield{doi}%%
    \clearfield{location}%%
    \clearfield{place}%%
        \clearfield{number}
    }
}
\AtBeginBibliography{\small}

\newcommand{\identity}{\mathds{1}}

\allowdisplaybreaks

\usepackage[section]{placeins} % force figures into the same section they're placed in
\usepackage{float} % supplies H placement option

\newlength{\topgap}
\newlength{\bottomgap}
\def\dividingline{\\[\topgap]&\makebox{\color{lightgray}\rule{2in}{0.5pt}}\\[\bottomgap]}

\title{\bf Tools for Mathematical Ludology}

\author{
    Paul Riggins$^1$ and David McPherson$^2$
    \medskip \\
    $^1$ \emph{Berkeley Center for Theoretical Physics, Department of Physics,} \\
    \emph{University of California, Berkeley, CA 94720, USA} \medskip\\
    $^2$ \emph{Department of Electrical Engineering and Computer Sciences,} \\
    \emph{University of California, Berkeley, CA 94720, USA} \medskip\\
    Email: priggins@berkeley.edu, david.mcpherson@eecs.berkeley.edu
}

\date{}

\begin{document}

\maketitle

\begin{abstract}
    We propose the study of mathematical ludology, which aims to formally interrogate questions of interest to game studies and game design in particular.
    The goal is to extend our mathematical understanding of complex games beyond decision-making---the typical focus of game theory and artificial intelligence efforts---to explore other aspects such as game mechanics, structure, relationships between games, and connections between game rules and user-interfaces, as well as exploring related gameplay phenomena and typical player behavior.
    In this paper, we build a basic foundation for this line of study by developing a hierarchy of game descriptions, mathematical formalism to compactly describe complex discrete games, and equivalence relations on the space of game systems.
\end{abstract}

\section{Introduction}
\label{sec:Introduction}

The art of game design has advanced tremendously in recent years, fueling and being fueled by a modern renaissance in tabletop and video gaming.
The methods and vocabulary used by designers and presented in game design schools has become increasingly sophisticated, as efforts have been made to systematize our understanding of game design elements and process (e.g., \cite{tekinbas_rules_2003,koster_theory_2005,adams_game_2012,schell_art_2019,engelstein_building_2019}).
Games are designed and used to entertain, train, educate, tell stories, sell products, study psychology, and simulate war, to name a few of their roles.
They are also advancing as experimental tools, probing real-world questions through the design and play of publicly accessible board and video games \cite{reddie_next-generation_2018}.

However, little mathematical attention has been given to questions of game design.
Games are distinct among artistic media in that the interactive systems underlying them can often be precisely defined.
This has enabled great mathematical progress in understanding game-centric decision-making processes, through efforts in game theory and artificial intelligence (AI) (e.g., \cite{yannakakis_artificial_2018}), but this progress has been largely isolated from other, ``softer'' subfields of game studies and design \cite{melcer_games_2015,melcer_toward_2017}.
This definability of games, however, could also be used to formally explore questions of interest to those softer subfields:
What makes a game broken, or balanced, or hyper-competitive?
How much does a game mechanic matter for overall gameplay?
How are these two games related?
What's the best user interface to reflect the underlying rules?
What makes this game hard to learn?
Can we predict behavior or strategy in one game based on data from a similar game?

Designers and players can have strong intuitions for questions like these and others, and those intuitions often draw upon the precisely definable underlying game systems.
This is an opportunity to complement existing game design discourse with mathematically formal underpinnings.
We thus propose the study of \emph{``Mathematical Ludology'': a mathematical exploration of the space of games and their properties in order to better understand game design principles, gameplay phenomena, and player behavior in complex games.}
The goal of the present paper is to lay groundwork for this exploration by developing some basic perspectives and accompanying mathematical framework through which complex games can be precisely and compactly described, and their design interrogated.

% We intend particularly to emphasize analytical methods in this line of study, rather than simulation, as our primary tool for describing games.
% Some progress has already been made in developing simulation-based computational metrics for game design features \cite{browne_evolutionary_2011}, and such methods may be indispensable for certain questions---simulations will at least be useful as a source of experimental data.
% AI agents do not necessarily play like humans \cite{khalifa_modifying_2016}, however, and results based on simulation can be difficult to explain.
% Our goal in mathematical ludology is especially to discover how much we can describe games and even predict their gameplay before players sit down to play, using analytical tools inspired from the observation of other games---perhaps even enabling back-of-the-envelope approximations usable by ludologers or designers on the fly.

Mathematical ludology will be unavoidably experimental at some level, since we are chiefly interested in human interactions with games.
Observational data, such as human gameplay data, will be important for developing and testing analytical models.
Simulations using AI agents may provide useful approximations \cite{khalifa_modifying_2016} of human play---indeed, some progress has already been made on simulation-based measurements of game design properties \cite{browne_evolutionary_2011}.
Ultimately, our goal is to discover how much we can describe games and predict their gameplay before players sit down to play, using analytical methods inspired from such measurements---perhaps even enabling back-of-the-envelope approximations usable by ludologers or designers on the fly.
In this sense, mathematical ludology can be viewed as an experimental science of complex games, using real-world observations and formal methods to develop descriptive and predictive models of games.

We think there is great potential benefit to this formal exploration, even beyond game design.
Identifying important mechanics or inter-game relationships could be useful for training general gameplaying AI agents, or developing approximate game theoretic solutions.
Learning to derive a user interface from logical game rules could be useful in varied applications of interactive system design.
Understanding facets of typical player behavior in complex games could help interpret gameplay data from simulation games (e.g., \cite{reddie_next-generation_2018}), making it more possible to isolate laboratory effects (behavior unique to gameplaying) from behavior that reflects real-world responses.

By the same token, mathematical ludology will necessarily be a multidisciplinary effort (like most of game studies), and will benefit from existing research in game theory, general gameplaying, cognitive science, psychology, interactive system design, wargaming, and other fields, not to mention game design.
Two especially notable companion fields are game design efforts towards ``game grammar,'' building formal methods to atomize and diagram games (e.g., \cite{cousins_elementary_2004,koster_grammar_2005,adams_game_2012,stephane_game_2006}); and recent computational efforts in digital archaeoludology \cite{browne_modern_2018}, building on prior work in automated game design \cite{browne_evolutionary_2011}.
These are the two research areas which, to our knowledge, are most similarly interested in mathematical questions of game design and structure, and we will discuss them more later on.

The rest of this paper proceeds as follows: 
In \cref{sec:GameDescriptions}, we describe a hierarchy of game descriptions that will help frame and contextualize different questions of interest.
In \cref{sec:UnderlyingGameSystem}, we define our lowest level formal description of a discrete game system, which will serve as a foundation for further study.
We also discuss why existing formal descriptions are insufficient for our purposes.
We discuss how to build and interpret game trees and automata in \cref{sec:GameTreesAndAutomata}.
Using these game trees, we develop equivalence relations on the space of game systems in \cref{sec:AgencyEquivalence}, with \cref{sec:EquivalenceTechnicalDefinitions} containing the technical details.
\cref{sec:StructuredNotation} builds on the formal description of \cref{sec:UnderlyingGameSystem}, offering more powerful notation to ease the expression and analysis of complex games.
We discuss the relationship of our work to formal methods in game design and digital archaeoludology in \cref{sec:RelatedWork}, and wrap up with discussion in \cref{sec:Discussion}.

\section{Game Description Hierarchy}
\label{sec:GameDescriptions}

\begin{figure}
    \centering
    \includegraphics[scale=1.00]{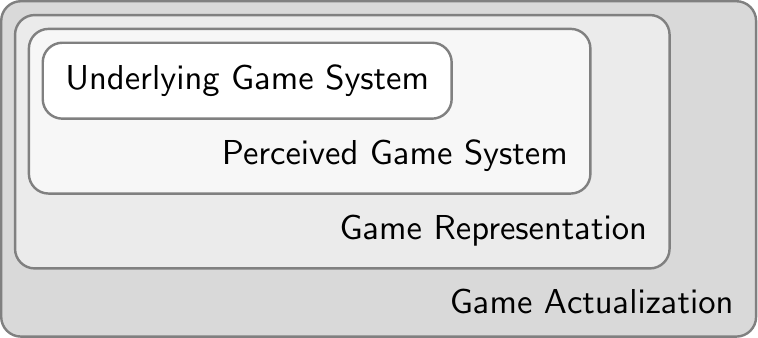}
    \caption{The game description hierarchy.
        We focus on Underlying Game Systems in this paper.
    } 
    \label{fig:GameDescriptionHierarchy}
\end{figure}

In order to isolate some of the facets that make up a game, and contextualize the work of this paper, we propose a four-tier hierarchy of game descriptions.
We will illustrate these tiers by contrasting the following two games:
standard chess, played by two people moving black and white pieces on a checkered board;
and blindfold chess, played by two people dictating their moves (e.g., ``knight to c3''), and each having to remember the game state, without any external record of it.

At the lowest tier, there is the \emph{underlying game system} (UGS), or simply \emph{game system} when there is no risk of confusion.
This will be the focus of the present paper.
A UGS describes the underlying rules and mechanics of the game,\footnote{Comparable to the ``constituative rules'' of \cite{tekinbas_rules_2003}.} the logic of a game, without any specification of the user interface or the information given to (or hidden from) the players.
Specifically, a UGS provides:
\begin{itemize}[noitemsep]
  \item Who is playing the game
  \item A way to describe the current game state
  \item How the game may be set up at the beginning
  \item The choices each player may make at each state
  \item The consequences of those choices, which alter the game state
  \item The results of a finished game, like who won or lost, or the final score
\end{itemize}
Standard and blindfold chess can be described with identical underlying game systems, since they share the same logical rules.

The second tier is a \emph{perceived game system} (PGS).
This builds on a UGS by adding an information layer, specifying what players are told (and not told) about the UGS and its state, other players' information, and how player knowledge might interact with the UGS rules (e.g., to describe how players obtain new information, or to describe epistemic games \cite{thielscher_gdl-iii:_2017}).
A PGS is needed to account for hidden game states (e.g., a secret hand of cards) or hidden game rules (e.g., \emph{Betrayal at House on the Hill} \cite{glassco_betrayal_2004}, the card game Mao, or \emph{Gloomhaven} \cite{childres_gloomhaven_2017}).
In a PGS for a perfect information game, like standard chess, every aspect of the UGS and its state are common knowledge among the players. 
In blindfold chess, the UGS rules, initial state, and each move are common knowledge, but the game state is otherwise hidden.
We find this information layer useful to separate from the UGS for two reasons: first, because the UGS is interesting to study in its own right, and second, because information specification can be very recursive and complicated (e.g., ``I know that you know that I don't know whether we're playing this game or that game''), and it is useful to have a separate object to ground this specification.
We offer some brief suggestions on how hidden information might be treated in \cref{sec:HiddenInformation}, for instance by employing intervals on the space of UGSs.

Note it is useful to distinguish between \emph{information} given to players, which is explicitly revealed through the game, and \emph{knowledge} that players have, which may additionally include memories or inferences.
For instance, the current game state in blindfold chess is never given as information (after the initial state), but it may be part of player knowledge (if each player can remember how it has evolved).
A perceived game system specifies information given to players, while inference and memory are left to player models (see below).

The third tier description is a \emph{game representation}.
A game representation builds on a PGS by additionally specifying the game's user interface, which should respect the rules and hidden information described in the PGS.
Standard chess uses pieces on a board as its interface, while blindfold chess uses a purely audio interface with certain accepted utterances.
It is useful to separate the interface from earlier tiers, because two game representations might have identical PGSs and yet very different interfaces (e.g., versions of tic-tac-toe, see \cref{ex:ArithmeticTic-tac-toe}), or identical interfaces but very different PGSs or UGSs (e.g., Go and Gomoku).
This interface specification may need varying levels of fidelity or abstraction for different research questions.
It may suffice to describe a graph resembling the board shape, for instance, with symbols that can be placed on graph nodes to represent pieces on squares, but for understanding various gameplay effects it might also be useful to distinguish piece shapes, card iconography, where players are sitting \cite{rogers_6_2019}, etc.

And finally, the fourth tier is a \emph{game actualization}.
This has all the elements of a game representation, while also including means of communicating to each player the appropriate information, providing a realization of the user interface, and providing means of actually playing the game.
Game actualizations are the ``real-world games'' that people actually play and interact with.
A physical chess set with a rulebook could be an actualization of standard chess, while just a rulebook might suffice for blindfold chess; an appropriate piece of software could also function as an actualization for either.

In contrast to game theoretic game descriptions, we do not include player preferences or payoffs in the description of the game at any of these levels.
We relegate these and everything else about each player into a \emph{player model}, including their preferences, skill level, play style, faculties of memory and inference, personal familiarity with the game, grudges against other players, or anything else that might be relevant for the question of interest.
These will be key to studying player behaviors and gameplay phenomena, and will likely benefit from existing research and methods in other fields, like game theory and cognitive science.

\section{Underlying Game Systems}
\label{sec:UnderlyingGameSystem}

The focus of the present paper will be on mathematizing underlying game systems and investigating some of their properties.
Specifically, we will focus on games that are discrete in time and space and that have a finite state space; \cref{def:UnderlyingGameSystem} can describe the game system of any such game.
This captures the majority of board and card games, and many video games.
It can approximate real-time or continuous-space games, but only insofar as they can be discretized into a finite state space.
A different formalism will be needed to describe improvisational games, like tabletop role-playing games or nomic games, where game rules are created and modified in unforeseeable ways during the course of play.\footnote{Such a formalism will at least need an alternative form of state space, perhaps related to suggestions in \cite{hammond_schumpeterian_2007}.}

The formalism (\cref{def:UnderlyingGameSystem}) we develop in this paper is not the only way to describe a game system, but we capture here those minimal elements essential to the logic of a game, while also making clear what agency each player has versus what is outside their control.
It will provide our foundation on which to begin formalizing games:

\begin{definition}
    \label{def:UnderlyingGameSystem}
    An \emph{(underlying) game system} $\mathcal{G}$ with $n$ players is a $9$-tuple $\mathcal{G} = \langle \mathcal{P}, \mathcal{T}, \mathcal{S}_0, \mathcal{D}, \mathcal{A}, C, L, \mathcal{O}, \Omega \rangle$, where:
  \begin{itemize}
    \item $\mathcal{P} = (1, \ldots, n)$ is a list of \emph{players}, agents which may make decisions in the game.
    \item $\mathcal{T} = (T_1, \ldots, T_m)$ is a finite list of finite sets, called \emph{substate tracks}. The set of \emph{game states} $\mathbb{S}$ is given by $\mathbb{S} = T_1 \times \cdots \times T_m.$
    \item $\mathcal{S}_0 \subset \mathbb{S}$ is a set of \emph{initial conditions}.
    \item $\mathcal{D}$ is a set of \emph{decisions}, the choices which players may make in order to influence (but not directly change) the game state. This is extended with the \emph{null decision} $0 \notin \mathcal{D}$ to form $\mathcal{D}_0 \equiv \mathcal{D}\cup\{0\}$.
    \item $\mathcal{A}$ is a set of \emph{actions} $a: \mathbb{S} \to \mathbb{S}$, which can directly modify the game state. 
    \item $C$ is a \emph{consequence function} $C(d_0^n, s)$ which takes a decision tuple $d_0^n\in\mathcal{D}_0^n$ (i.e., one possibly null decision per player) and state $s\in\mathbb{S}$ and returns a nonempty set of \emph{consequences}: a set of pairs $(p_a, a)$, where $p_a \in (0,1]$ is a non-zero probability and $a$ is an action or product of actions. The sum of probabilities in the set must equal 1.
        These are the consequences of decisions, which may be outside any one player's control.
    \item $L: \mathcal{P} \times \mathcal{D}\to 2^{\mathbb{S}}$ is a \emph{legality function}, which returns a (possibly empty) subset of $\mathbb{S}$ for each player $p\in\mathcal{P}$ and decision $d\in \mathcal{D}$, reflecting when that player can make that decision. 
        The \emph{legal set of decisions for player $p$ at state $s$} is the set $L_p(s) \equiv \{ d\in\mathcal{D}: s \in L(p,d) \}$.
        A decision $d \in L_p(s)$ is \emph{legal} for $p$ at $s$, and \emph{illegal} otherwise.
    \item $\mathcal{O}$ is the set of \emph{outcomes} that can result from the game.
    \item $\Omega$ is an \emph{outcome function} $\Omega: \mathcal{S}_\text{ter}\to\mathcal{O}$, where $\mathcal{S}_\text{ter} \equiv \{ s\in\mathbb{S}: L_p(s) = \varnothing \text{ for all }p\in\mathcal{P}\}$ is the set of \emph{terminal game states}.
        Intuitively, $\mathcal{S}_\text{ter}$ is the set of game states at which no legal decisions can be made, so the game ends and the result is computed by $\Omega$.
  \end{itemize}
The collection of all such objects $\mathcal{G}$ forms the space of game systems $\mathbb{G}$.
\end{definition}

\cref{def:UnderlyingGameSystem} provides all the elements necessary for a game system (see bulleted list in \cref{sec:GameDescriptions}): players ($\mathcal{P}$), the game state space ($\mathbb{S}$, factorized via $\mathcal{T}$), initial game states ($\mathcal{S}_0$), decisions available to players ($\mathcal{D}$), when those decisions are legal ($L$), the possible consequences of those decisions ($C$), how the game state is changed as a result ($\mathcal{A}$), and possible game outcomes ($\mathcal{O}, \Omega$).
It allows for mixed sequential and simultaneous play, deterministic or nondeterministic, and makes no additional assumptions about the content of games, not even presuming the existence of boards and pieces.

The separation of decisions (player choices) from actions (changes to the game state) via consequence functions provides a formal separation between what individual players can and cannot control.
Consequence functions are necessary to handle random chance and simultaneous play.
In both cases, each player can influence play by their chosen decision, but the ultimate effect on the game state (the consequent action) is determined probabilistically and/or after considering the simultaneous decisions of other players (see \cref{def:GameplayAlgorithm,ex:FlippingCoin}).
In sequential, deterministic games, it is possible to have a one-to-one mapping between decisions and actions---individual players can directly determine game state changes---and so consequence functions are superfluous (see \cref{ex:GuessingGame}).

Note we have included randomness in the game description itself in order to allow a total conceptual separation of game system and player models---in contrast to typical game theory or general gameplaying formalisms, which relegate randomness to an extra fictional player who behaves randomly \cite{rasmusen_games_2007,thielscher_general_2010,piette_ludii_2019}.
(There are also other reasons we have not used one of these existing formal game descriptions, see \cref{sec:ExistingDescriptions}.)

The following algorithm can be used (by players) to play any game with an underlying game system described by \cref{def:UnderlyingGameSystem}:
\begin{definition}[Gameplay Algorithm] \hfill
    \label{def:GameplayAlgorithm}
    \begin{enumerate}
        \item All players must agree on some $s_0\in\mathcal{S}_0$. Let the current state be $s' = s_0$.
        \item \label{item:GameplayAlgorithm-StartLoop} Each player $p$ must select one decision from their respective legal set $L_p(s')$ at the current state. If $L_p(s') = \varnothing$ for a player $p$, that player is assigned the null decision: $d_p = 0$.
        \item If $d_p = 0$ for all $p$, the game is over. Go to step \ref{item:GameplayAlgorithm-StopLoop}.
        \item Compute the set of consequences from the decision tuple and the current state: $c = C((d_1, \ldots, d_n), s') = \{(p_1, a_1), \ldots, (p_m, a_m)\}$.
            Randomly select a single consequent action from $c$, where $a_j$ is selected with probability $p_j$.
        \item Compute the new game state $s'' = a_j s'$. Repeat from step~\ref{item:GameplayAlgorithm-StartLoop}, with the new current state $s' = s''$.
        \item \label{item:GameplayAlgorithm-StopLoop} The current state is a terminal state, $s' \in \mathcal{S}_\text{ter}$. Compute the outcome $\Omega(s')$.
    \end{enumerate}
\end{definition}

This algorithm can be used to describe which game systems are \emph{playable}, and to distinguish between \emph{legal} and \emph{illegal} play (see \cref{sec:CompleteGamesAndLegalPlay}).
It is also closely related to the construction of game trees (see \cref{sec:GameTreesAndAutomata}).

\subsection{Basic Notation}
\label{sec:SomeNotation}

Before we proceed to examples, let us introduce some basic notation for use with \cref{def:UnderlyingGameSystem}.
Here and throughout, we choose to use fairly compact symbolic notation, for instance ``$L(\text{P1},\text{flip}) = (-)_\text{coin}$'' which means ``Player `P1' can legally choose decision `flip' when the track `coin' takes value `$-$'.''
This has the benefit of efficiency, but may take some practice to read and write fluently.
We include commentary with each example to build familiarity.
We will introduce additional structured notation in \cref{sec:StructuredNotation}, which will help to improve readability, compactness, and ease of analysis for more complex game descriptions.

\emph{Regarding game states.} 
We write $(v)_t$ to express that track $T_t$ takes value $v$.

These may be combined using product, sum, and overline notation (for Boolean AND, OR, and NOT, respectively) to express a subset (or ``\emph{slice}'') of state space $\mathbb{S}$.
For instance: $u = (1)_a(2)_b + \overline{(3)}_c$ is the subset of $\mathbb{S}$ such that (($T_a$ takes value 1 AND $T_b$ takes value 2) OR ($T_c$ does NOT take value 3)).

\emph{Regarding actions.} 
By $a: S\mapsto (v_1)_1\cdots(v_k)_k$, we mean that for any state in the slice $S \subset \mathbb{S}$, the action $a$ changes the value of track $T_1$ to $v_1$, and so on to track $T_k$, acting as the identity on any tracks not appearing on the right-hand side.
It also acts as the identity on any state $s\notin S$.
E.g., if $a: \mathbb{S}\mapsto (1)_a$ and we have some state $s = (2)_a(3)_b$, then $a\cdot s = (1)_a(3)_b$.

\emph{Regarding outcome functions.}
We take $\Omega: S\mapsto \omega$ to mean $\Omega(z) = \omega$ for all terminal states $z \in S$.

\subsection{Basic Examples}
\label{sec:BasicExamples}

Here are two examples of very simple games.
Their respective game trees are illustrated in \cref{fig:CoinFlipGameTree,fig:GuessingGameTree}.
We will see examples of more complicated games after introducing richer notation in \cref{sec:StructuredNotation}.

\begin{example}[Flipping a coin]
    \label{ex:FlippingCoin}
    Two players face off and flip a coin once.
    The first player wins on heads, the second on tails.
\begin{align*}
    &\mathcal{P} = \{ \text{P1}, \text{P2} \} \\
    &\mathcal{T}: T_\text{coin} = \{ -, \text{heads}, \text{tails} \} \\
    &\mathcal{S}_0 = (-)_\text{coin} \dividingline
    &\mathcal{D} = \{ \text{flip} \} \\
    &L(\text{P1}, \text{flip}) = L(\text{P2}, \text{flip}) = (-)_\text{coin} \\
    &C((\text{flip}, \text{flip})) = \{ (1/2, A_\text{heads}), (1/2, A_\text{tails}) \} \\
    &\mathcal{A} = \{ A_\text{heads}, A_\text{tails} \}, \quad A_c: (-)_\text{coin} \mapsto (c)_\text{coin} \dividingline
    &\mathcal{O} = \{ \text{P1 win}, \text{P2 win} \} \\
    &\Omega: (\text{heads})_\text{coin} \mapsto \text{P1 win}, \quad (\text{tails})_\text{coin} \mapsto \text{P2 win}
\end{align*}

The lines roughly divide the setup, gameplay, and ending portions of the system.
From the initial condition, both players can only legally choose the decision ``flip.''
As a consequence of this joint decision, the value of track $T_\text{coin}$ becomes heads or tails with a 50-50 chance.
The game then ends, with no more legal decisions available, and the winner is declared.

Remember at this point that all of the symbols used for players, decisions, actions, etc. (e.g., ``P1'', ``flip'', or ``$A_\text{coin}$'') are only labels---we could have just as well used pictographs.
We treat these labels as semantically void so far as the game system is concerned (see coordinates in \cref{def:StructuredNotationSamples} for contrast), though they may certainly become relevant as part of a game representation, where aesthetic description and theming may matter for the user interface.
\end{example}

\begin{example}[Guessing game]
    \label{ex:GuessingGame}
    One player chooses a number between one and five.
    A second player tries to guess the number, winning if they guess it correctly and losing otherwise.
\begin{align*}
    &\mathcal{P} = \{ \text{P1}, \text{P2} \}  \\
    &\mathcal{T}: T_{\text{P1}} = T_{\text{P2}} =\{ -, 1, 2, 3, 4, 5 \}\ \\
    &\phantom{\mathcal{T}:}\ T_{\text{picked}} = T_{\text{guessed}} = \{ \text{no}, \text{yes} \}\ \\
    &\mathcal{S}_0 = (-)_{\text{P1}}(-)_{\text{P2}} (\text{no})_\text{picked} (\text{no})_\text{guessed} \dividingline
    &\mathcal{D} \sim \mathcal{A}\quad (C: \text{trivial}) \\
    &\mathcal{A} = \{ \text{1}_{\text{P1}}, \text{1}_{\text{P2}}, \ldots, \text{5}_\text{P1}, \text{5}_\text{P2} \}, \\
    &\qquad \qquad i_\text{P1}: \mathbb{S}\mapsto (i)_\text{P1}(\text{yes})_\text{picked} \\
    &\qquad \qquad i_\text{P2}: \mathbb{S}\mapsto (i)_\text{P2}(\text{yes})_\text{guessed} \\
    &L(\text{P1}, i_\text{P1}) = (\text{no})_\text{picked} \\
    &L(\text{P2}, i_\text{P2}) = (\text{yes})_\text{picked} (\text{no})_\text{guessed} \\
    &L(p, d) = \varnothing,\quad \text{otherwise} \dividingline
    &\mathcal{O} = \{ \text{P2 win}, \text{P2 lose} \} \\
    &\Omega: (1)_\text{P1}(1)_\text{P2} + \cdots + (5)_\text{P1} (5)_\text{P2} \mapsto \text{P2 win},  \\
    &\phantom{\Omega: }\ \text{otherwise} \mapsto \text{P2 lose}
\end{align*}

This is a sequential, deterministic game, with a one-to-one mapping between decisions and actions (denoted $\mathcal{D} \sim \mathcal{A}$).
Thus we only need to list out the actions, instead of the decisions as well.
The consequence function just maps each decision to its corresponding action with probability 1, regardless of state; we denote this with $(C: \text{trivial})$.

In this game P1 picks a number by choosing one of the decisions $\{ 1_\text{P1},\ldots,5_\text{P1} \}$, while P2 is forced to pick the null decision because no legal options are available.
Then P2 guesses a number by choosing a decision $\{ 1_\text{P2},\ldots,5_\text{P2} \}$, while P1 picks the null decision.
The game then ends, since no one can legally make another decision.

This is a prime example where the underlying game system is not sufficient to capture the real world experience.
In particular, this game is only interesting if P2 can't see P1's guess.
A perceived game system would specify this hidden information, in particular hiding $T_\text{P1}$ from P2 while still letting them see $T_\text{picked}$ so they know when to guess.
The underlying game system is only concerned with the game logic below this information layer.

Alternatively, an equivalent underlying game system could do away with $T_\text{picked}$ and $T_\text{guessed}$ altogether, replacing the legality function mappings with $L(\text{P1}, i_\text{P1}) = (-)_\text{P1}$ and $L(\text{P2}, i_\text{P2}) = \overline{(-)}_\text{P1}(-)_\text{P2}$.
A corresponding perceived game system would then selectively reveal to P2 whether the current state $s \in (-)_\text{P1}$ or not, instead.
We say more about what we mean by ``equivalent'' in \cref{sec:AgencyEquivalence}.
\end{example}

\subsection{Hidden Information}
\label{sec:HiddenInformation}

We will not treat hidden information carefully here, but here are some brief suggestions on how it could be approached for perceived game systems built on \cref{def:UnderlyingGameSystem} as the underlying game system.

Many games only hide aspects of the current state from the players, for instance the cards in an opponents hand, or which number player one has guessed.
Players otherwise have full common knowledge of the rules and game system.
Such hidden state information can be specified simply by flagging which track values are visible or invisible to each player, as in the comments following \cref{ex:GuessingGame}.
These flags could even be included as additional tracks in $\mathcal{T}$ as part of the game state.
(This is essentially how Ludii manages hidden information \cite{piette_ludii_2019}.)

However, it is also very possible for other aspects of the game system to be hidden.
This is necessary to model story-based games, like many video games or modern campaign or legacy board games (e.g., \cite{childres_gloomhaven_2017}), where future scenarios and rules are unknown until they are reached.
It is also important for modeling and understanding the dynamics of tutorial games: games where the rules are revealed to players a bit at a time, often as a method for teaching first-time players.

Games like these, with partially hidden systems, could be modeled by specifying intervals on the space of underlying game systems $\mathbb{G}$, representing all of the possible games that each player is told they might be playing.
This would also require appropriate maps from each of these possible games (and their states, and the information available to other players) to the true game (and its actual state, and the actual information available to other players).

\subsection{Complete Games and Legal Play}
\label{sec:CompleteGamesAndLegalPlay}

The gameplay algorithm \cref{def:GameplayAlgorithm} provides our first technical distinction between kinds of gameplay:
Any gameplay which strictly follows the gameplay algorithm is said to be \emph{legal play}, and a game state is said to be \emph{legally accessible} if it can be produced in some application of this algorithm.
Gameplay which deviates from the algorithm constitutes \emph{illegal play}: for instance, starting the game at some $s\notin\mathcal{S}_0$, neglecting the legality function when choosing decisions, ignoring the probabilities when selecting a consequent action, modifying the game state in a way not given by a consequent action, etc.
It will be interesting in future work to explore the effects of illegal play, intentional and accidental---especially to understand the effects of accidental rule-breaking (common in some complex games, e.g., \emph{Mage Knight} \cite{chvatil_mage_2011}), and how to make the designer's goals of gameplay robust against it.
For the remainder of this paper, however, we will assume legal play.

The gameplay algorithm additionally furnishes a sense of when a game description is complete:
\begin{definition}
    \label{def:CompleteGame}
    We call a game system $\mathcal{G}$ \emph{complete} or \emph{playable} if the algorithm \cref{def:GameplayAlgorithm} can always be faithfully performed.
    That is, if $\mathcal{P}, \mathbb{S}, \mathcal{S}_0, \mathcal{D}, \mathcal{A}$ are all nonempty, the legality function $L$ is uniquely defined for all player-decision pairs, and the consequence function $C$ is uniquely defined at all legally accessible decision tuples ($d_0^n \in \mathcal{D}_0^n$) and states.
    If any terminal states are legally accessible, then we also require $\mathcal{O}$ to be nonempty and the outcome function $\Omega$ to be uniquely defined on all legally accessible terminal game states.
\end{definition}

It may in general be difficult to confirm if a game system is playable, because determining which game states are legally accessible is a nontrivial problem \cite{chinchalkar_upper_1996}.
The simplest way to ensure a game system is playable is to make it \emph{overcomplete}: with the consequence function $C$ uniquely defined for every decision tuple (except possibly the all-null tuple) and state, and the outcome function $\Omega$ uniquely defined on all terminal game states, in addition to $L$ uniquely defined on all player-decision pairs, and nonempty $\mathcal{P}, \mathbb{S}, \mathcal{S}_0, \mathcal{D}, \mathcal{A}, \mathcal{O}$.
In fact, such a game is guaranteed to play faithfully from any arbitrary state $s \in \mathbb{S}$, not just $s \in \mathcal{S}_0$, which may be interesting for exploring illegal play.

All examples in this paper are complete, though none of them are overcomplete. 
\cref{ex:FlippingCoin} does not define consequences for the tuple $(\text{flip},0)$, for instance.

\subsection{Other Game Formalisms}
\label{sec:ExistingDescriptions}

Let us now address why we have developed a new formal game description, rather than using an existing one---besides the issue of randomness mentioned above (including it in-system rather than as a fictional player).
The likely candidates are formal descriptions from game theory, general gameplaying, and formal methods in the game design community.

Game theoretic descriptions are too limited in the games they can practically express.
There are two issues in view here: first, game theoretic descriptions are unable to faithfully describe mixed sequential and simultaneous play.
Strategic- and extensive-form games respectively describe simultaneous and sequential games well, but lose important nuance when trying to mix the two \cite{cooper_communication_1992,salles_beyond_2008}.

Second, complex game descriptions are intractable with game theoretic formal descriptions.
Generally, either the full game can be written explicitly as a strategic, extensive, or combinatorial game (all intractable, e.g., for chess), or else game theorists rely on ad hoc natural language description and reader familiarity to communicate the rules before proceeding to analysis (e.g., \cite{beck_combinatorial_2008}). 
We want a way to formally, and tractably, describe game rules even for complex games.

We also find general gameplaying descriptions inadequate for our purposes.
GDL and its extensions \cite{love_general_2006,thielscher_general_2010,thielscher_gdl-iii:_2017} are currently able to express the widest variety of games, but through code that is often intractably verbose, and from which it is difficult to extract structural features of games \cite{piette_ludii_2019}.
\cref{def:UnderlyingGameSystem} bears some formal similarity GDL-II without hidden information, and we think is just as expressive, though we offer a simpler specification of state space, a different treatment of randomness, and prefer more compact and extensible notation.

More recent entrants on the field, RBG \cite{kowalski_regular_2019} and Ludii \cite{piette_ludii_2019} more closely reflect the goals of mathematical ludology, with Ludii in particular being concerned with questions of game structure and design (see \cref{sec:RelatedWork}).
However, both require the construction of a game board tightly integrated with the rules, distinguishing them as game representations (Tier 3 game descriptions).
The underlying game systems \cite{piette_ludii_2019,kowalski_regular_2019} are related to \cref{def:UnderlyingGameSystem} (Ludii more than RBG), but designed for use with particular software and associated code bases, and with particular classes of games in mind.
We desire something general and self-contained that we can develop to study any discrete game, with the flexibility to relate to such projects and software while not being beholden to them.

There have also been formal descriptions and diagramming methods developed within the game design community (e.g., \cite{koster_game_2015,adams_game_2012,stephane_game_2006}), but they are built mostly as tools for high level design or analyzing subsystems within games.
They are unable to compactly capture the full detail of a game's rules, as we require.
See \cref{sec:RelatedWork} for further discussions on these and Ludii as related work.

\section{Game Trees and Automata}
\label{sec:GameTreesAndAutomata}

\begin{figure}
    \centering
    \includegraphics[scale=1.0]{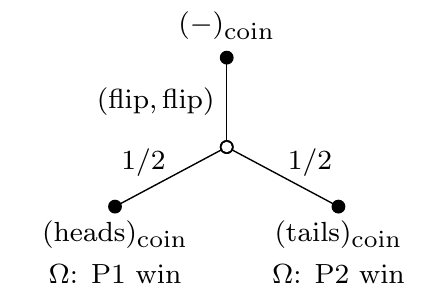}
    \caption{The game tree for the coin flipping game system in \cref{ex:FlippingCoin}, with all nodes and edges fully labeled: the root state node with state $(-)_\text{coin}$, the decision edge with tuple $(\text{flip},\text{flip})$, the chance edges both with probability $1/2$, and the terminal state nodes $(\text{heads})_\text{coin}$ (outcome: P1 win) and $(\text{tails})_\text{coin}$ (outcome: P2 win).
    Solid nodes are state nodes, while the open node is a chance node.
    }
    \label{fig:CoinFlipGameTree}
    \vspace{-7pt}
\end{figure}

\begin{figure}
    \centering
    \hspace{-3.5mm}
    \includegraphics[scale=0.95]{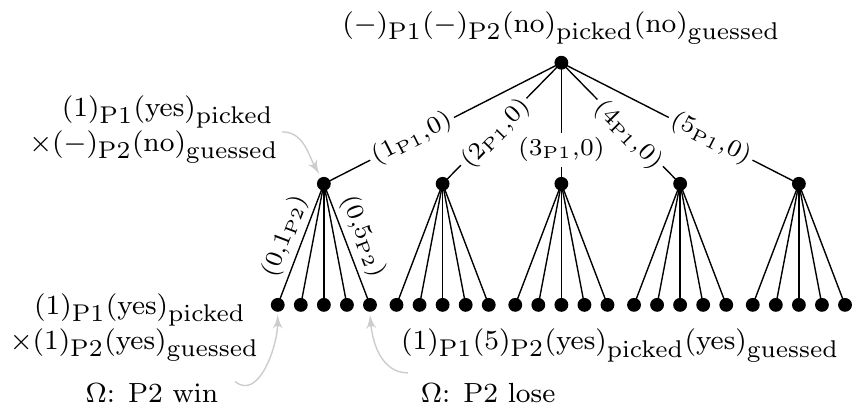}
    \caption{The game tree for the guessing game system in \cref{ex:GuessingGame}, with only a few state nodes and decision edges labeled, for brevity.
        Each of the five bottom subtrees have the same five decision tuples on their edges: $(0,1_\text{P2}), (0,2_\text{P2}),\ldots, (0,5_\text{P2})$.
        There are no chance nodes; this is a deterministic game.
    }
    \label{fig:GuessingGameTree}
    \vspace{-7pt}
\end{figure}

A complete game system from \cref{def:UnderlyingGameSystem} can be used to generate a (possibly infinite) game tree or a finite (possibly nondeterministic) game automaton.
These are useful for visualizing and analyzing the game systems, as well as for making connection with existing work in game theory and AI.
They are graphical representations of the algorithm \cref{def:GameplayAlgorithm}: each playthrough from that algorithm can be identified as a path from an initial node to a terminal node in one of these objects.
We will focus on game trees, rather than automata, in this paper.
Examples are illustrated in \cref{fig:CoinFlipGameTree,fig:GuessingGameTree}, and also \cref{fig:MatchingDecisionTreesExample,fig:BookkeepingReductionExample,fig:SinglePlayerReductionExample} (in \cref{sec:EquivalenceTechnicalDefinitions}).

\begin{definition}
    \label{def:GameSystemTree}
    The \emph{game trees} $\tau(\mathcal{G})$ of a complete game system $\mathcal{G}$ is a set of game trees $\tau(\mathcal{G}) \equiv \{ \tau(\mathcal{G},s_0): s_0\in\mathcal{S}_0 \}$, one for each initial condition. 
    Each game tree $\tau(\mathcal{G},s_0)$ is given by the following construction:
  \begin{enumerate}
    \item Draw a root node, assigned the initial state $s_0$.
    \item \label{item:GenerateGameTreeAlgorithm-StartLoop}
        For each terminal node $w$ in the current tree, with assigned state $s(w)$ but no assigned outcome, do the following:
    \begin{enumerate}
        \item 
            If $s(w) \in \mathcal{S}_\text{ter}$, assign the outcome $\Omega(s(w))$ to the terminal node $w$, then stop for node $w$. If $s(w)\notin\mathcal{S}_\text{ter}$, then proceed:
        \item Generate the set of all legal decision tuples at this state from the legal set $L_p(s)$ for each player: $D_0^n(s(w)) \equiv \{ (d_1, \ldots, d_n): d_p = 0 \text{ if } L_p(s(w)) = \varnothing,\text{ else }d_p\in L_p(s(w)) \}$
        \item \label{step:DecisionMatrix} For each tuple $d_0^n \in D_0^n(s(w))$, draw a new child node $w'$ with a directed edge from $w$ to $w'$.
            Assign $d_0^n$ to this edge.
        \item For each child node $w'$ of $w$, do the following:
            \begin{enumerate}
                \item Compute the set of consequences $c' = C(d_0^n,s(w))$. 
                \item If $|c'| = 1$, i.e. $c' = \{ (1, a) \}$, assign the state $a\cdot s(w)$ to $w'$.
                    Otherwise:
                \item For each probability-action pair $(p_i,a_i) \in c'$, draw a new child node $w''$ with a directed edge from $w'$ to $w''$.
                    Assign $p_i$ to this edge, and the state $a_i\cdot s(w)$ to $w''$.
            \end{enumerate}
    \end{enumerate}
\item If all terminal nodes in the current tree have outcomes assigned to them, stop: this tree $\tau(\mathcal{G},s_0)$ is finished. Otherwise, repeat from step~\ref{item:GenerateGameTreeAlgorithm-StartLoop} with the current tree.
  \end{enumerate}
\end{definition}

\captionsetup[figure]{skip=9pt}

In summary, each resulting tree has the following structure:
\emph{State nodes} have assigned states and outgoing \emph{decision edges}, which have assigned decision tuples.
These decision edges lead to either new state nodes, or to unlabeled \emph{chance nodes} which have outgoing \emph{chance edges} labeled with probabilities.
These chance edges lead to new state nodes.
State nodes may be further subdivided into \emph{single-player nodes}, in which only a single player has legal decisions available; \emph{multiplayer nodes}, in which multiple players have legal decisions available; and \emph{terminal state notes}, which correspond to terminal states and have outcomes assigned to them.
Single-player nodes can be said to \emph{belong} to the appropriate player.

It is worth noting that these game trees are not guaranteed to be finite, even if the game in practice would end in finite time.
Consider a game of rock-paper-scissors, for instance, where the two players can in principle keep tying forever.
In cases like these, it might be more practical to consider game automata instead of game trees.
These would be similarly defined, except each game state would only appear once.
If a newly drawn edge would lead to a state that exists elsewhere in the automaton, it would be directed to the existing state node instead of drawing a new one.
Because the state space always has a finite size by virtue of \cref{def:UnderlyingGameSystem}, and any non-determinism follows specific probabilities, the game automaton would be most similar to a probabilistic automaton \cite{rabin_probabilistic_1963}.\footnote{
    The 4-tuple $\langle \mathcal{D}, \mathcal{A}, C, L \rangle \subset \mathcal{G}$, which determines the moment-to-moment gameplay rules, could be combined into a single stochastic transition function $\delta(s,d_0^n)$ that takes a state and decision tuple and returns all possible subsequent states with corresponding probabilities.
(Illegal transitions would have probability 0.)
A probabilistic automaton could use this transition function, with state space $\mathbb{S}$ and input alphabet $\mathcal{D}_0^n$.
}

A game tree generated by \cref{def:GameSystemTree} is close to the classic game theoretic extensive form game, for instance as given in \cite{rasmusen_games_2007}, except without any information sets.
In this sense, a game system might almost be considered a grammar to generate extensive form games without hidden information.
The notable exception is how we handle simultaneous play.
Unlike extensive form games, some nodes in the tree (so-called multiplayer nodes) require multiple players to make a decision at the same time.
This avoids the ambiguities inherent in the information set construction of simultaneous play in extensive form games \cite{bonanno_set-theoretic_1992}, which relies on sequential moves with hidden information, and can be experimentally different from true simultaneous play \cite{cooper_communication_1992,salles_beyond_2008}.

\captionsetup[figure]{skip=9pt}

With this difference, it would be more accurate to consider a game tree by our definition as a hybrid between game theoretic extensive form and strategic form games.
A game tree with no multiplayer nodes is simply an extensive form game, in which a single player has control of each node and chooses one of the outgoing edges to follow, ultimately producing an outcome at a terminal node.
In a game tree with multiplayer nodes, however, each multiplayer node acts as a strategic form game: several players must simultaneously make a decision, which is then evaluated by an umpire to choose an outgoing edge to follow.
(We describe this strategic form game with a \emph{decision matrix}, see \cref{def:DecisionMatrix}.)
With this understanding, we could use our game tree \cref{def:GameSystemTree} along with a single marker as a user interface in a game representation, for playing any game without hidden information.

\section{Agency Equivalence}
\label{sec:AgencyEquivalence}

The game system \cref{def:UnderlyingGameSystem} is deliberately flexible, to act as a foundation for a wide variety of game descriptions.
However, it's flexibility means that there are several ways to express the same game system.
A key goal in mathematical ludology will be to measure distances between different games, so here we take a first step: what does it mean for two game systems to be equivalent?
We will develop two important senses of this, \emph{game tree equivalence (up to relabeling)} and \emph{agency equivalence}.
We describe them heuristically here, leaving the technical details for \cref{sec:EquivalenceTechnicalDefinitions}.

Game tree equivalence up to relabeling (\cref{def:GameTreeEquivalenceUpToRelabeling}) matches game systems if they produce the same game trees, with some differences in aesthetic labeling---e.g., the names of decisions or states can differ, but the probabilities cannot. 
Two such game systems will be very similar in their formal descriptions.

Agency equivalence (\cref{def:AgencyEquivalence}) matches game systems if they offer players the same agency, that is, if the systems offer players the same sorts of meaningful choices with the same sorts of consequences.
We will define this by performing a series of reductions on game trees to prune spurious differences, declaring two game systems equivalent if their reduced trees match.
There are four main kinds of difference that we consider spurious for this purpose: bookkeeping subtrees, single-player subtrees, symmetry-redundant subtrees, and decision matrix redundancies.
Recall throughout that we are talking about underlying game systems (Tier 1 game descriptions), so differences due to hidden information and user interfaces are not in view right now.

A \emph{bookkeeping subtree} (\cref{def:BookkeepingSubtree}) is a portion of a game tree where there is only one decision available at each state.
Though there may be randomness involved, play continues on the subtree inevitably, without any chance for player influence.
Perhaps a game has a cleanup phase, for instance, where players must discard all cards and end their turn.
One game system may lump these together, while another might include a state where players must discard, followed by another state where they must end their turn.
We do not consider these different from the standpoint of player agency.

A \emph{single-player subtree} (\cref{def:SinglePlayerSubtree}) is a portion of the game tree where the same single player makes several deterministic decisions in a row.
There is no difference in options or outcomes if the player makes these decisions one at a time or all at once.
Consider pawn promotion in chess.
One game system might have a player choose to advance a pawn to the final row, then from that state choose what piece to promote it into.
Another might include move-and-promote-into-queen, move-and-promote-into-knight, etc., all as lumped decisions with no intermediate state.
We do not consider these different from the standpoint of player agency.

A \emph{symmetry-redundant subtree} (\cref{def:SymmetryRedundantSubtree}) is a portion of the game tree that is unnecessary because it duplicates a sibling subtree.
Consider the first move in a game of tic-tac-toe.
Although there are technically nine options, there are in practice only three: center, corner, or side.
Playing on the right side does produce a different game state than playing on the left side, but because of the symmetry of the game board the substance of the remaining decisions is identical.
If players were restricted to only play in the center, left-middle side, or bottom-left corner on the first turn, we would consider this identical to tic-tac-toe from the standpoint of meaningful player agency.

Finally, a \emph{decision matrix redundancy} (\cref{def:DecisionMatrixRedundancy}) occurs when a player has two decisions at a state that would have identical results---it really does not matter which one they pick.
This player would have the same agency if they had only one of those decisions available.

Putting these together:
\begin{definition}
    \label{def:AgencyEquivalence}
    We say two game systems $\mathcal{G}$ and $\mathcal{G}'$ are \emph{agency equivalent} if their respective game trees can be made equivalent up to relabeling (\ref{def:GameTreeEquivalenceUpToRelabeling}) by performing the following reductions, as many times as necessary, in any order:
    \begin{itemize}[noitemsep]
        \item Bookkeeping subtree reduction (\ref{def:BookkeepingSubtree})
        \item Single-player subtree reduction (\ref{def:SinglePlayerSubtree})
        \item Symmetry-redundant subtree reduction (\ref{def:SymmetryRedundantSubtree})
        \item Decision matrix redundancy reduction (\ref{def:DecisionMatrixRedundancy})
    \end{itemize}
\end{definition}

Note this definition can only be usefully applied for finite game trees, though infinite trees could be truncated at a certain depth and similarly compared.

Practically speaking, establishing equivalence will probably be easiest by transforming and subsequently comparing the game system grammars, not the game trees.
This definition gives intuitive and technical guidance on what those transformations must accomplish.

\section{Structured Notation}
\label{sec:StructuredNotation}

The game system \cref{def:UnderlyingGameSystem} is sufficient for describing the game system of any finite discrete game, but writing down complex games with only the basic notation of \cref{def:UnderlyingGameSystem} and \cref{sec:SomeNotation} could be intractably verbose.
The basic notation also does little to expose game mechanics or related rules that might appear in many games, like the $n$-in-a-row victory condition of Tic-tac-toe and Connect Four, or the sliding movement of a queen in Chess and Amazons.
Relatedly, the labels given to tracks, decisions, etc. are semantically void so far as the formal description is concerned, being only convenient monikers to the ludologer writing them; integer labels need not respect addition or multiplication on the integers, for instance.

In order to express games in a form most appropriate for their analysis, it will be helpful to use \emph{structured notation}. 
By this we simply mean notation for expressing game systems (\cref{def:UnderlyingGameSystem}) besides or in addition to the basic notation introduced earlier, especially which emphasizes structural patterns within game descriptions.
We expect such structured notations will be critical for developing distance metrics on the space of game systems $\mathbb{G}$, for instance, since human intuitions for game similarity tend to rely strongly on high-level patterns in game designs.
In this paper we use formal grammar-like notations, but different structured notations like diagrammatic methods or the ludeme-tree code of Ludii could be useful in other contexts (see \cref{sec:RelatedWork}).

So without further ado, let us introduce a few key additional features that will ease the expression and analysis of a wide variety of games:
\begin{definition}[Sample structured notations]
    \label{def:StructuredNotationSamples}
    The following notational features may be used along with the basic notation of \cref{def:UnderlyingGameSystem} and \cref{sec:SomeNotation}, to express game systems: 
    \begin{enumerate}
        \item A set of \emph{ending states} $E\subset\mathbb{S}$ may be provided, at which the game should end even if legal decisions would otherwise be available.

            \emph{Partial legality functions} $\hat{L}_i(p, d)$ may be defined on subsets of $\mathcal{P}\times\mathcal{D}$ (returning subsets of $\mathbb{S}$), and a single player-decision pair $(p, d)$ may appear in multiple partial legality functions $\hat{L}_1(p, d), \ldots, \hat{L}_m(p, d)$.
            Each $\hat{L}$ adds additional restrictions on the legality of $(p,d)$.

            The overall legality function for $(p, d)$ is then given by $L(p, d) = \hat{L}_1(p, d) \cap \cdots \cap \hat{L}_m(p, d) \cap \overline{E}$.

        \item Each element of $\mathcal{P}, \mathcal{T}, \mathcal{D}, \mathcal{A}, \mathcal{O}$, as well as each substate value $v\in T\in\mathcal{T}$, may be assigned one or more 
            \begin{enumerate}
                \item \emph{coordinates} $\phi$: an element of some mathematical object $\Phi$, with all the structure of that object available for manipulations on $\phi$.
                    For instance, $\Phi$ may be a group or graph and $\phi$ may be a group element or node of a graph, so that group addition or node adjacency are well-defined notions inherited from $\Phi$ that can be used in the game description.
                \item \emph{tags} $t$: a label used to group related elements.
                    Those elements of the same type (players, substate tracks, track values, decisions, actions, outcomes) which share the same tag may be referred to collectively by the tag name.
                    E.g., $(t)^\text{p}$ is a set of players which share the same tag $t$, and similarly $(t)^\text{t}, (t)^\text{v}, (t)^\text{d}, (t)^\text{a}, (t)^\text{o}$ for the other types, respectively.
            \end{enumerate}

        \item Any element of the game description may be defined through the use of auxiliary \emph{ludemic functions}.
            These may take as arguments any element of the game description or current game state, including associated tags and coordinates, and return whatever is useful (e.g., state slices, a choice from a list, \ldots).
    \end{enumerate}
\end{definition}

Each of these features captures and highlights a different kind of structure in a game description.
Ludemic functions are particularly flexible, intended to capture atomic patterns in a game's design
(see \cref{sec:RelatedWork}), like the movement pattern of a chess piece or the probability distribution of an opposed dice roll in \emph{Risk} \cite{lamorisse_risk:_1959}.
They may be defined for an individual game description, but will be most useful when collected in a catalog, for portable use in different games that share those same patterns of game logic.
Using ludemic functions from this catalog may improve readability and analysis of game descriptions, but with the trade-off of requiring reader familiarity with each function used.\footnote{An extreme example of this is Ludii, in which every aspect of the game description must be drawn from a library of ``ludemes'' \cite{soemers_ludii_2019}, a kind of computational cousin to a catalog of ludemic functions (see \cref{sec:RelatedWork} for further discussion).
    This can result in compact and readable descriptions for a variety of games \cite{noauthor_notitle_nodate}, but requires outside knowledge of every ludeme used.
}

Functions in this catalog may be related to each other: for instance, a chess queen's movement function is a special case of a 2D translation, which is a special case of an $n$D translation.
It even relates to the $n$-in-a-row pattern found in tic-tac-toe, since there must be $n$-in-a-row blank spaces for a queen to move $n+1$ spaces.
Such relationships could be used to build distance metrics on the catalog of ludemic functions, which would then be useful in building metrics on the space of game systems.
We leave this catalog and related metrics to future work, though we illustrate a couple examples of ludemic functions in \cref{ex:RockPaperScissors,ex:ArithmeticTic-tac-toe}.

Additional kinds of structured notation will likely be useful for different classes of games.
For instance, some games involve a stack or some other kind of finite ``memory'' of past states (e.g., \emph{Magic: The Gathering} \cite{garfield_magic:_1999}, or chess stalemate rules).
Similarly, some games like chess employ ``foresight,'' determining the legality of a move now based on what moves might become legal as a result (e.g., the ``check'' mechanic).
Defining a special kind of ``memory'' substate alongside the usual substate tracks, or a ``foresight'' legality function, might be useful for such games.

We emphasize that not every kind of structure is admissible here.
Any game with an unbounded memory requirements would still not be possible, for instance, because that would require an unbounded state space.
Describing games of that sort would require more fundamental modifications of \cref{def:UnderlyingGameSystem}.

\subsection{Boolean brackets}
\label{sec:BooleanBrackets}

Before we proceed to examples, let us see how we can use these new features to express subsets of state space, adding to our structured notation. Note we will use $\phi(e)$ to refer to the coordinate of a game element $e$, and $T(s)$ to refer to the value of track $T$ at state $s$.

We can describe state slices by leveraging Boolean relations on coordinates and tags.
We delineate these Boolean calculations by square brackets.
For instance, $[\phi(T) > 5]$ is either $\mathbb{S}$ if the track $T$ has integer coordinate greater than 5, or $\varnothing$ otherwise.
If an element has multiple coordinates, we denote the intended by subscripts, for instance $[\phi_i(T) > 5]$, $[\phi(T) >_i 5]$, or $[\phi(T) > 5]_i$ for the $i$th coordinate.
(Recall that tracks can have coordinates independent from their values.)

To extend this to track values, we let $[B(v(T))] \equiv \{ s \mid B(T(s)) \}$ for Boolean relation $B$ and state $s \in \mathbb{S}$.
For instance, $[v(T) \in (b)^\text{v}] = \{ s \mid T(s) \in (b)^\text{v} \} \subset \mathbb{S}$ is the set of all states where the track $T$ takes some value that has tag $b$.
These will usually return nonempty proper subsets of $\mathbb{S}$.

We might call this bracket notation ``Boolean slice functions'' or ``Boolean brackets.''
They could be considered a special case of ludemic functions, though it may be useful to conceptually distinguish them when considering atomic elements of games (see \cref{sec:RelatedWork}).
We'll illustrate Boolean brackets along with other new notation through the following examples.

\subsection{Examples with Structured Notation}
\label{sec:StructuredNotationAndExamples}

We now present three examples---Rock-Paper-Scissors, Tic-Tac-Toe, and the hand game Chopsticks---to illustrate this structured notation, describing each to explain new shorthand and build familiarity with interpreting the formalism.

\begin{example}[Rock-Paper-Scissors]
    \label{ex:RockPaperScissors}
    Two players play rock-paper-scissors; best two out of three wins.
\begin{align*}
    &\mathcal{P} = (\text{P1},\text{P2}), \quad \mathcal{O} = \mathcal{P} \\
    &\mathcal{T}: \quad  \mathcal{P} \leftarrow [0,2] \\
    &\mathcal{S}_0 = (0)_\text{P1} (0)_\text{P2}  \dividingline
    &\mathcal{D} = \{ \text{rock}, \text{paper}, \text{scissors} \} \sim \mathbb{Z}_3 \\
    &\mathcal{A}: \quad \mathcal{P} \ni p: \mathbb{S} \mapsto (+1)_p;\quad  \identity: \mathbb{S}\mapsto\mathbb{S} \\
    &C: (d_1, d_2)\mapsto \text{RPS}((d_1, d_2), \identity, \text{P1}, \text{P2}) \\
    &\hat{L}(p, d) = \mathbb{S} \dividingline
    &E = E_\text{P1}\cup E_\text{P2}, \quad E_p = (2)_p, \quad p \in \mathcal{T} \\
    &\Omega: E_p \mapsto p
\end{align*}
Let's unpack the notation a bit:

There are no tags in this example.
We reuse the list $\mathcal{P}$ as track names, so $\mathcal{P} \leftarrow [0,2]$ means we have two tracks, P1 and P2, which each take the values $[0,2] = \{0,1,2\}$.
We implicitly take these values to have integer coordinates---on the ring $\mathbb{Z}$, with the usual addition and multiplication---inferred from the integer interval notation.
(We will use the interval notation $[a, b]$ to refer to integers by default, rather than reals.)

The decisions, by contrast, have coordinates on $\mathbb{Z}_3$ (the ring of integers modulo 3) with $\phi(\text{rock}) = 0$, $\phi(\text{paper}) = 1$, and $\phi(\text{scissors}) = 2$, and respecting modular arithmetic (e.g., $1-2 = -1 \text{ mod } 3 = 2$).
Note we use $ \sim $ to associate coordinates to sets, drawing correspondence from the written or canonical ordering of each.

Two of the actions are labeled by the players, P1 and P2, and increment the corresponding track by 1.
Note we've leveraged a coordinate operator as a nice shorthand: $a: \mathbb{S}\mapsto (+1)_T$ is shorthand (see also \cref{sec:SomeNotation}) for $a: s\mapsto(\phi(T(s)) + 1)_T, s\in\mathbb{S}$.
The third action $\identity$ is the identity function.

The consequence function uses the ludemic function $\text{RPS}((d_1,d_2),a_0,a_1,a_2)$.
This takes a tuple of two decisions with coordinates on $\mathbb{Z}_3$,\footnote{Since the null decision hasn't been given a coordinate on $\mathbb{Z}_3$, the RPS function is only well-defined on non-null decisions in this example.} and returns the action $a_i$ if $\phi(d_1) - \phi(d_2) = i$.
Note here we've effectively written the consequence function as $C: d_0^n \mapsto a$.
This is shorthand for $C(d_0^n,s) = \{ (1,a) \}$ for all $s \in \mathbb{S}$.

The partial legality function $\hat{L}$ adds no restrictions: any decision is legal for any player at any state that isn't an ending state ($s \notin E$).

The ending states are those states where either track takes value 2---which track determines the outcome (winner), though we have been lazy and defined the outcomes as $\mathcal{O} = \{\text{P1}, \text{P2}\}$ instead of the more descriptive $\{\text{P1 wins}, \text{P2 wins}\}$.
In either case, it's up to the players to determine how they value each outcome, according to their player models.

It is interesting to note that the outcome function $\Omega$ is not uniquely defined on all terminal game states: the state $z = (2)_\text{P1}(2)_\text{P2}$ is a terminal state (since $z \in E$), but since $z \in E_\text{P1}$ \emph{and} $z \in E_\text{P2}$,  it is ambiguous which outcome results.
The state $z$ is not legally accessible, however, so the game system is still complete and playable.
If this ambiguity were not present, and if $C$ were also defined on the partially null tuples $(d,0)$ and $(0,d)$, this system would be overcomplete.
\end{example}

\begin{example}[Arithmetic Tic-tac-toe]
    \label{ex:ArithmeticTic-tac-toe}
Two players take turns claiming numbers from the set $\{ -4, -3, -2, -1, 0, 1, 2, 3, 4 \}$.
Each number can only be chosen once, and a player wins by gathering any three numbers that can add to 0.
The game ends when either a player wins or all numbers have been claimed, at which point the game ends in a draw if nobody has a triple that sums to 0.
\begin{align*}
    &\mathcal{P} = (\text{P1},\text{P2}), \quad \mathcal{O} = \mathcal{P} \cup \{ \text{draw} \} \\
    &\mathcal{T}:  \begin{array}[t]{lllll}
        \text{(numbers)} &    [-4,4] &\leftarrow& \mathcal{P} \cup \{-\} & \\
                         &   \text{turn}       &\leftarrow& \mathcal{P} &\sim \mathbb{Z}_2
    \end{array} \\
    &\mathcal{S}_0 = (-)_\text{(numbers)} (\text{P1})_\text{turn} \dividingline
    &\mathcal{D} \sim \mathcal{A}\quad (C: \text{trivial}) \\
    &\mathcal{A} = \mathcal{P}\times\text{(numbers)} \ni (p, n): \mathbb{S} \mapsto (p)_n (+1)_\text{turn} \\
    &\hat{L}(p, (p',n)) = (-)_n (p)_\text{turn}  [p = p'] \dividingline
    &E = E_\text{P1}\cup E_\text{P2},  \\
    &\quad E_p = \text{TripleSumsToZero}(\text{(numbers)}, p) \\
    &\Omega: E_p \mapsto p,\quad  \mathcal{S}_\text{ter} \setminus E \mapsto \text{draw}
\end{align*}
This is our first example with tags, and coordinates on track values, so let's again unpack this a bit.

There are ten tracks here, which in basic notation we might write $T_{-4}, T_{-3}, \ldots, T_4, T_\text{turn}$.
The first nine tracks have integer names, and also (implicitly by means of the integer interval notation) corresponding integer coordinates.
They are grouped into the tag set $(\text{numbers})^\text{t}$, and each track takes the values $\{ \text{P1}, \text{P2}, - \}$.
The tenth track has name ``turn,'' and its values have coordinates on $\mathbb{Z}_2$: $\phi(\text{P1}) = 0$, $\phi(\text{P2}) = 1$.

We use the tags to compactly express the initial condition.
In general, we can reference a set of tracks to define a slice: e.g., if tracks $T_1, T_2$ both have tag $t$ and possible value $v$, then $(v)_{(t)} = (v)_1(v)_2 \subset \mathbb{S}$.

There are eighteen actions, named as pairs with one member $p$ from $\mathcal{P}$ and one member $n$ from $(\text{numbers})$.
As we have here, we will often drop the type superscript on tag sets where there is no risk of confusion, just writing $(\text{numbers})$ instead of $(\text{numbers})^\text{t}$.
Each action marks the number $n$ as claimed by player $p$, and toggles the turn.

Each player $p$ can legally claim number $n$ for player $p'$ (i.e., make decision $(p',n)$) if the number $n$ isn't taken, it is $p$'s turn, and they are claiming it for themselves ($[p=p']$).

The ending states are defined via the ludemic function $\text{TripleSumsToZero}(K, v)$, which takes a set $K$ of tracks $t$ with integer coordinates $\phi(t)\in\mathbb{Z}$ and possible value $v \in t$, and returns the union of all slices $(v)_{t_1}(v)_{t_2}(v)_{t_3}$ such that $\phi(t_1) + \phi(t_2) + \phi(t_3) = 0$.
E.g., $\text{TripleSumsToZero}(\text{(numbers)}^\text{t}, \text{P1}) = (\text{P1})_{-4}(\text{P1})_{0}(\text{P1})_{4} + (\text{P1})_{-4}(\text{P1})_{1}(\text{P1})_{3} + \cdots$.

Note that this is essentially the same game system we would use to describe tic-tac-toe:
\captionsetup[figure]{skip=8pt}
\begin{figure}[H]
    \centering
    \begin{tabular}{|r|r|r|}\hline
        $-3$ & $2$  & $1$ \\\hline
        $4$  & $0$  & $-4$ \\\hline
        $-1$ & $-2$ & $3$  \\\hline
    \end{tabular}
    \caption{``Magic square'' correspondence between the arithmetic and grid-based versions of tic-tac-toe.}
    \vspace{-8pt}
\end{figure}
\noindent To replicate the more familiar grid-based thinking, we could replace the integer coordinates $[-4,4]$ of the nine (numbers)$^\text{t}$ tracks with 2D coordinates $[1,3]^2\subset\mathbb{Z}^2$, and replace the ending states with an analogous function $E_p = \text{ThreeInARow}(\text{(numbers)}, p)$, for instance using the definition of $n$-in-a-row on $\mathbb{Z}^d$ from \cite{beck_combinatorial_2008}.
These two game systems would be game tree equivalent up to relabeling, and with identical players and outcomes.
The version with tracks on $\mathbb{Z}^2$ might be easier to compare to a 4x4 version of tic-tac-toe (e.g., extended with tracks $[1,4]^2$ and everything else the same), but the arithmetic version might be easier to compare to a version with extended tracks $[-5,5]$, which has no clear grid-based analog.
Which structured expression is more useful will depend on the question being asked.

This is also interesting case study in technical game atoms (see \cref{sec:RelatedWork}):  here we see how different game descriptions might use different ludemic functions, reflecting different atomic game concepts, and yet describe the same game system.
\end{example}

    \def\dashdividingline{\\[\topgap]&\makebox{\color{lightgray}\tikz\draw[dashed] (0,0) -- +(1.5in,0);}\\[\bottomgap]}
\begin{example}[Chopsticks]
    \label{ex:Chopsticks}
    \hspace{8mm} From Wikipedia: ``Chopsticks is a hand game for two players in which players extend a number of fingers from each hand and transfer those scores by taking turns to tap one hand against another'' \cite{wikipedia_contributors_chopsticks_2019}.
    A full natural language rules description can be found there, and is implemented and explained here:
\begin{align*}
    &\mathcal{P} = (\text{P1},\text{P2}), \quad \mathcal{O} = \mathcal{P} \\
    &\mathcal{T}:  \begin{array}[t]{lllll}
        \text{(hands, P1)} & \{ \text{1L}, \text{1R} \} &  \leftarrow [0, 4] &  \sim \mathbb{Z}_5, [0, 4] \\
        \text{(hands, P2)} & \{ \text{2L}, \text{2R} \} &  \leftarrow [0, 4] &  \sim \mathbb{Z}_5, [0, 4] \\
        & \text{turn}              &\leftarrow \mathcal{P} &\sim \mathbb{Z}_2
    \end{array} \\
    &\mathcal{S}_0 = (1)_\text{(hands)} (\text{P1})_\text{turn} \dividingline
    &\mathcal{D} \sim \mathcal{A}\quad (C: \text{trivial}) \\
    &\mathcal{A} = (\text{add})^\text{a}\cup(\text{transfer})^\text{a} \dashdividingline
    &\hat{L}_1(p, \mathcal{D}) = (p)_\text{turn} \dashdividingline 
    &(\text{add})^\text{a} =  \text{(hands)}\times\text{(hands)} \ni \\
    &\quad \quad \quad (h, g): (v)_h (v')_g \mapsto (v' +_1 v)_g (+1)_\text{turn} \\
    &\hat{L}_2(p, (h,g)) = [h \in (\text{hands}, p)]\, \overline{(0)}_h \overline{(0)}_g  \dashdividingline 
    &(\text{transfer})^\text{a} = \text{(hands)}\times\text{(hands)}\times[1, 4] \ni \\
    &\hspace{5mm} (h, g, n): (v)_h (v')_g \mapsto (v -_2 n)_h (v' +_2 n)_g (+1)_\text{turn} \\
    &\hat{L}_2(p, (h,g,n)) = [h,g \in (\text{hands}, p)]\ [v(h) > n]_2  \\
    &\quad \times [v(g) + n < 5]_2\ [v(h) - v(g) \neq n]_2 \dividingline
    &\Omega: (0)_{(\text{P1})}\mapsto \text{P2}, \quad (0)_{(\text{P2})}\mapsto \text{P1}
\end{align*}
This is a more involved example, with a more mathematical game, to illustrate more uses of coordinates.
It also illustrates how our notation can compactly express some complex games even without any ludemic functions (though with ample use of Boolean brackets).

Here we have five tracks, named 1L, 1R, 2L, 2R, (reflecting players' hands, with five fingers each) and turn. 
We have three tags: hands, P1, and P2.
In general, we will notate the intersection of two tag-sets $(t)^\text{y}, (t')^\text{y}$ of the same type y by the comma-separated $(t, t')^\text{y}$, so we see for instance that the track 1L has two tags: hands and 1.

The turn track has values $\{\text{P1}, \text{P2}\}$, and coordinates on $\mathbb{Z}_2$, a cyclic turn track like in \cref{ex:ArithmeticTic-tac-toe}.
This is a turn-based game, because the partial legality function $\hat{L}_1$ prevents any player from acting out of turn, and every action advances the turn track.

The (hands)$^\text{t}$ tracks have integer values $[0,4]$, but because we see the $\sim$ coordinate assignment we will \emph{not} implicitly assume these have just integer coordinates: instead we see they have two sets of coordinates, first $\mathbb{Z}_5$ (coordinate 1) and second $[0,4]\subset\mathbb{Z}$ (coordinate 2).
The order matters here, because we will need to refer to them later on.

We have grouped the actions with tags for readability, but notice how each uses the coordinates:
The (add) actions allow a player to tap a hand $h$ against another hand $g$, using modular arithmetic (coord. 1) to add the number on $h$ to the number on $g$.
Players can legally tap one of their own not-empty hands against any other non-empty hand, as long as it's their turn.

The (transfer) actions allow a player to tap a hand $h$ against another hand $g$, using integer arithmetic (coord. 2) to transfer 1--4 points from $h$ to $g$.
Players can legally transfer between their own hands on their turn, as long as neither hand ends up empty (with zero or five fingers: $[v(h) > n]_2, [v(g) + n < 5]_2$) and the hands don't just switch values ($[v(h) - v(g) \neq n]_2$).

The game starts with 1 finger on each hand, and ends when one player has both hands empty, with the other player winning.
Notice we don't need to specify additional ending states (we can take $E = \varnothing$), because the partial legality functions are already sufficient: no legal decisions can be made when it reaches the turn of a player with two empty hands.
\end{example}

\section{Related work}
\label{sec:RelatedWork}

As mentioned in the Introduction, there are many other lines of research that may be relevant to progress in mathematical ludology. 
Here we discuss the two which, to our knowledge, share the most particular kinship.

The first is an effort within the game design community, launched especially by Benjamin Cousins \cite{cousins_elementary_2004} and Raph Koster \cite{koster_grammar_2005}, to formally break down games into their constituent atoms and develop notation to diagram them, as a means to improve the critical study and design of games.
This has produced several notational systems and notions of game atoms, each best suited to unpacking different kinds of features from different kinds of board games and video games (e.g., \cite{cousins_elementary_2004,koster_theory_2005,stephane_game_2006,adams_game_2012,koster_game_2015}).
Most of these notations have been developed by game designers for game designers, and as such prioritize higher-level concepts in order to better survey, iterate, and improve game designs.
Even the most detailed of these notational systems (``machination diagrams'' \cite{adams_game_2012}) would struggle to provide the level of rule detail needed to build a game tree, like we require from \cref{def:UnderlyingGameSystem} for careful analysis.

It may be interesting to build structured notations (beyond \cref{def:StructuredNotationSamples}) which leverage these game diagramming methods, in order to probe higher-level analysis questions and visualizations.
It will also will be interesting to formally understand and build technical analogues (e.g., ludemic functions) for the various game atoms and constituents that have been proposed by designers before and after Cousins and Koster: choice molecules \cite{tekinbas_rules_2003}, primary elements \cite{cousins_elementary_2004}, ludemes \cite{koster_theory_2005,koster_atomic_2012,parlett_whats_2016,browne_evolutionary_2011}, mechanisms (e.g., as compiled in \cite{engelstein_building_2019}), and others.

Another, more recent research direction is digital archaeoludology, spearheaded by the Digital Ludeme Project (DLP) \cite{browne_modern_2018}.
The DLP aims to bring computational attention and AI methods to historical studies of traditional strategy games.
As a key part of this, games are modeled and their designs studied in Ludii, a software program under development which provides a means to digitally describe and play games \cite{piette_ludii_2019}.

The game properties and relationships of particular interest to the DLP, e.g., strategic depth or phylogenetic distance \cite{browne_modern_2018}, are interesting case studies in our more general interest in game metrics, and will need to overcome similar challenges. 
For instance, the same game can be described many ways in Ludii code, and we think this flexibility should not be allowed to unduly affect phylogenetic distance measurements when they are developed---similar to our considerations around agency equivalence, and an issue for any game comparison method.

Ludii is intended as a general-purpose tool with AI capabilities \cite{piette_ludii_2019,noauthor_notitle_nodate} and we may find it useful, e.g., for game description testing, gameplay data collection, and automating certain game measurements (especially via simulations).
Ludii has chosen a particular game atom for its descriptions, the ludeme,%
\footnote{The ludeme of Ludii \cite{soemers_ludii_2019}, and its predecessor Ludi \cite{browne_evolutionary_2011}, is rather different from the ludeme of Koster \cite{koster_theory_2005}, and includes both smaller and larger structures: every part of a Ludii description is a Ludii-ludeme, from a solitary Boolean AND function to the entire game itself \cite{soemers_ludii_2019}. 
    See \cite{parlett_whats_2016} for discussion of the origins and pre-Ludi(i) uses of ``ludeme,'' or \cite{koster_atomic_2012,koster_theory_2005} for an early understanding of Koster's ludeme.
}
somewhat related to ludemic functions and bringing similar benefits and challenges (see \cref{sec:StructuredNotation}).
It will be interesting to explore its meaning and use alongside other candidate game atoms.
Ludii will also be interesting as a case study for Tiers 3 and 4 of our game description hierarchy.
In our terms, Ludii provides game representations and actualizations (see \cref{sec:GameDescriptions}), since hidden information and interface specification are woven into Ludii game descriptions, and the software provides means of playing the described games \cite{piette_ludii_2019}.
We will be well-positioned to study the structure of Ludii game descriptions after studying the more primitive underlying and perceived game systems (Tiers 1 and 2), like we have begun in this paper.
For instance, Ludii uses a tree structure to arrange the atoms of its game descriptions \cite{browne_modern_2018,piette_ludii_2019}, a kind of structured notation which our approach could enable us to generalize beyond Ludii code, and to probe its uses and limitations.

\section{Discussion}
\label{sec:Discussion}

We have proposed a new line of game studies research, mathematical ludology, 
which aims to formally explore the space of games and their properties in order to better understand game design principles, gameplay phenomena, and player behavior in complex games.
In this paper we have developed some basic mathematical formalism that allows the compact description of complex, finite discrete games, with notation that helps to expose their structure for later comparison and analysis.
We have focused mostly on underlying game systems, the first tier of our game description hierarchy: much work still remains to flesh out the specification of hidden information and user interfaces.
Existing progress in game theory and general gameplaying may be helpful for some of those next steps, since these efforts have typically begun at the second tier and higher.

We have also begun developing equivalence relations on the space of game systems, a first step in learning to formally compare games.
It will be interesting next to develop a variety of distance metrics on the space of games, each highlighting different aspects of game descriptions.
This will require a careful understanding of game structure, as well as an understanding of which differences are not meaningful, in the same way that many game system differences were not meaningful in establishing agency equivalence.
It will be useful to develop distance metrics on the space of ludemic functions, ludemes, or other technical game atoms, as part of this effort.
Taken together, these metrics will enable formal taxonomies of games, and provide a reference point for comparing gameplay trends or strategies in related games.

A natural next step will be to consider the distinct properties of common structures in game rules, for instance game phases, resources, or decks as common (and perhaps uniquely identifiable) kinds of substates in underlying game systems.
This could be a stepping stone towards comparing games, building user interfaces, or generating natural language descriptions of rules.
For instance, we have a method under development to derive user interface graphs from underlying game systems, which may benefit from understanding such structures \cite{riggins_in_progress}.

Great progress has already been made in the study, design, and application of games, thanks to the efforts of academics and practitioners in both mathematical and non-mathematical disciplines.
As game studies continues to be a multidisciplinary effort, we hope that progress in mathematical ludology might help to bridge some of the gap between mathematical and game design experts, enriching the work of both.

\section*{Acknowledgments}

We would like to thank
Stephen Crane,
Marquita Ellis,
Ryan Janish, 
Kiran Lakkaraju,
Kweku Opoku-Agyemang, 
Stephen Phillips, 
and
Ben Wormleighton
for useful discussions.
We would also like to thank Vlaada Chv\'atil for designing the board game \emph{Mage Knight} \cite{chvatil_mage_2011}, which inspired us towards this line of research.

\renewcommand{\bibname}{References}
\printbibliography

@manual{soemers_ludii_2019,
	title = {Ludii User Guide},
	url = {https://www.ludii.games/LudiiUserGuide-0.2.0.pdf},
	pagetotal = {55},
	version = {0.2.0},
	organization = {Maastricht University},
	author = {Soemers, Dennis J. N. J. and Piette, Éric and Stephenson, Matthew and Browne, Cameron},
	date = {2019-08-12},
    urldate = {2019-11-10},
}

@incollection{salles_beyond_2008,
	location = {Berlin, Heidelberg},
	title = {Beyond Normal Form Invariance: First Mover Advantage in Two-Stage Games with or without Predictable Cheap Talk},
	isbn = {978-3-540-79831-6 978-3-540-79832-3},
	url = {http://link.springer.com/10.1007/978-3-540-79832-3_12},
	shorttitle = {Beyond Normal Form Invariance},
	abstract = {Von Neumann (1928) not only introduced a fairly general version of the extensive form game concept. He also hypothesized that only the normal form was relevant to rational play. Yet even in Battle of the Sexes, this hypothesis seems contradicted by players’ actual behaviour in experiments. Here a reﬁned Nash equilibrium is proposed for games where one player moves ﬁrst, and the only other player moves second without knowing the ﬁrst move. The reﬁnement relies on a tacit understanding of the only credible and straightforward perfect Bayesian equilibrium in a corresponding game allowing a predictable direct form of cheap talk.},
	pages = {215--233},
	booktitle = {Rational Choice and Social Welfare},
	publisher = {Springer Berlin Heidelberg},
	author = {Hammond, Peter J.},
	% editor = {Pattanaik, Prasanta K. and Tadenuma, Koichi and Xu, Yongsheng and Yoshihara, Naoki},
	% editorb = {Salles, M. and Pattanaik, P. K. and Suzumura, K.},
	% editorbtype = {redactor},
	urldate = {2019-11-13},
	date = {2008},
	langid = {english},
	doi = {10.1007/978-3-540-79832-3_12},
}

@book{beck_combinatorial_2008,
	location = {Cambridge ; New York},
	title = {Combinatorial games: tic-tac-toe theory},
	isbn = {978-0-521-46100-9},
	series = {Encyclopedia of mathematics and its applications},
	shorttitle = {Combinatorial games},
	pagetotal = {732},
	pages = {53},
	number = {v. 114},
	publisher = {Cambridge University Press},
	author = {Beck, József},
	date = {2008},
	langid = {english},
	% note = {{OCLC}: ocn175284055},
	keywords = {Combinatorial analysis, Game theory},
}

@article{piette_ludii_2019,
	title = {Ludii - The Ludemic General Game System},
	url = {http://arxiv.org/abs/1905.05013},
	abstract = {While current General Game Playing ({GGP}) systems facilitate useful research in Artificial Intelligence ({AI}) for game-playing, they are often somewhat specialized and computationally efficient. In this paper, we describe an initial version of a "ludemic" general game system called Ludii, which has the potential to provide an efficient tool for {AI} researchers as well game designers, historians, educators and practitioners in related fields. Ludii defines games as structures of ludemes, i.e. high-level, easily understandable game concepts. We establish the foundations of Ludii by outlining its main benefits: generality, extensibility, understandability and efficiency. Experimentally, Ludii outperforms one of the most efficient Game Description Language ({GDL}) reasoners, based on a propositional network, for all available games in the Tiltyard {GGP} repository.},
	% journaltitle = {{arXiv}:1905.05013 [cs]},
	author = {Piette, Éric and Soemers, Dennis J. N. J. and Stephenson, Matthew and Sironi, Chiara F. and Winands, Mark H. M. and Browne, Cameron},
	urldate = {2019-11-13},
	date = {2019-05-16},
	eprinttype = {arxiv},
	eprint = {1905.05013},
    primaryClass = {cs},
	keywords = {Computer Science - Artificial Intelligence},
}

@book{browne_evolutionary_2011,
	location = {London},
	title = {Evolutionary game design},
	isbn = {978-1-4471-2178-7 978-1-4471-2179-4},
	series = {Springer-briefs in computer science},
	pagetotal = {122},
	publisher = {Springer},
	author = {Browne, Cameron},
	date = {2011},
	langid = {english},
	% note = {{OCLC}: 740624561},
}

@inproceedings{melcer_games_2015,
	location = {Pacific Grove, {CA}, {USA}},
	title = {Games Research Today: Analyzing the Academic Landscape 2000-2014},
	eventtitle = {International Conference on the Foundations of Digital Games},
	booktitle = {Proceedings of the 10th International Conference on the Foundations of Digital Games ({FDG})},
	author = {Melcer, Edward and Nguyen, Truong-Huy Dinh and Chen, Zhengxing and Canossa, Alessandro and El-Nasr, Magy Seif and Isbister, Katherine},
	date = {2015},
	langid = {english},
}

@article{bonanno_set-theoretic_1992,
	title = {Set-theoretic equivalence of extensive-form games},
	volume = {20},
	issn = {0020-7276, 1432-1270},
	url = {http://link.springer.com/10.1007/BF01271135},
	doi = {10.1007/BF01271135},
	pages = {429--447},
	number = {4},
	journaltitle = {International Journal of Game Theory},
	shortjournal = {Int J Game Theory},
	author = {Bonanno, G.},
	urldate = {2019-11-13},
	date = {1992-12},
	langid = {english},
}

@article{cooper_communication_1992,
	title = {Communication in Coordination Games},
	volume = {107},
	issn = {0033-5533, 1531-4650},
	url = {https://academic.oup.com/qje/article-lookup/doi/10.2307/2118488},
	doi = {10.2307/2118488},
	pages = {739--771},
	number = {2},
	journaltitle = {The Quarterly Journal of Economics},
	shortjournal = {The Quarterly Journal of Economics},
	author = {Cooper, R. and {DeJong}, D. V. and Forsythe, R. and Ross, T. W.},
	urldate = {2019-11-13},
	date = {1992-05-01},
	langid = {english}
}

@book{yannakakis_artificial_2018,
	location = {Cham, Switzerland},
	title = {Artificial Intelligence and Games},
	isbn = {978-3-319-63519-4 978-3-319-63518-7},
	pagetotal = {337},
	publisher = {Springer},
	author = {Yannakakis, Georgios N. and Togelius, Julian},
	date = {2018},
	% note = {{OCLC}: 1028525002},
}

@inproceedings{thielscher_general_2010,
	title = {A General Game Description Language for Incomplete Information Games},
	url = {http://dl.acm.org/citation.cfm?id=2898607.2898766},
	series = {{AAAI}'10},
	pages = {994--999},
	booktitle = {Proceedings of the Twenty-Fourth {AAAI} Conference on Artificial Intelligence},
	publisher = {{AAAI} Press},
	author = {Thielscher, Michael},
	date = {2010},
	% note = {event-place: Atlanta, Georgia},
}

@inproceedings{melcer_toward_2017,
	location = {Hyannis, {MA}},
	title = {Toward understanding disciplinary divides within games research},
	isbn = {978-1-4503-5319-9},
	url = {http://dl.acm.org/citation.cfm?doid=3102071.3106355},
	doi = {10.1145/3102071.3106355},
	eventtitle = {International Conference on the Foundations of Digital Games},
	booktitle = {Proceedings of the International Conference on the Foundations of Digital Games ({FDG})},
	publisher = {{ACM} Press},
	author = {Melcer, Edward and Isbister, Katherine},
	urldate = {2019-11-18},
	date = {2017},
	langid = {english},
}

@article{reddie_next-generation_2018,
	title = {Next-generation wargames},
	volume = {362},
	issn = {0036-8075, 1095-9203},
	url = {http://www.sciencemag.org/lookup/doi/10.1126/science.aav2135},
	doi = {10.1126/science.aav2135},
	pages = {1362--1364},
	number = {6421},
	journaltitle = {Science},
	shortjournal = {Science},
	author = {Reddie, Andrew W. and Goldblum, Bethany L. and Lakkaraju, Kiran and Reinhardt, Jason and Nacht, Michael and Epifanovskaya, Laura},
	urldate = {2019-11-20},
	date = {2018-12-21},
	langid = {english},
}

@inproceedings{thielscher_gdl-iii:_2017,
	title = {{GDL}-{III}: A Description Language for Epistemic General Game Playing},
	isbn = {978-0-9992411-0-3},
	url = {http://dl.acm.org/citation.cfm?id=3171642.3171823},
	series = {{IJCAI}'17},
	pages = {1276--1282},
	booktitle = {Proceedings of the 26th International Joint Conference on Artificial Intelligence},
	publisher = {{AAAI} Press},
	author = {Thielscher, Michael},
	date = {2017},
	% note = {event-place: Melbourne, Australia},
}

@report{love_general_2006,
	location = {Stanford University, Stanford, {CA}},
	title = {General Game Playing: Game Description Language Specification},
	url = {http://logic.stanford.edu/reports/LG-2006-01.pdf},
	number = {{LG}-2006-01},
	institution = {Stanford Logic Group, Computer Science Department},
	author = {Love, Nathaniel and Hinrichs, Timothy and Genesereth, Michael},
	date = {2006-04-04},
}

@article{kowalski_regular_2019,
	title = {Regular Boardgames},
	volume = {33},
	issn = {2374-3468, 2159-5399},
	url = {https://aaai.org/ojs/index.php/AAAI/article/view/3991},
	doi = {10.1609/aaai.v33i01.33011699},
	pages = {1699--1706},
	journaltitle = {Proceedings of the {AAAI} Conference on Artificial Intelligence},
	shortjournal = {{AAAI}},
	author = {Kowalski, Jakub and Mika, Maksymilian and Sutowicz, Jakub and Szykuła, Marek},
	urldate = {2019-11-20},
	date = {2019-07-17},
}

@book{rasmusen_games_2007,
	location = {Oxford and Malden, {MA}},
	edition = {4th ed},
	title = {Games and information: an introduction to game theory},
	isbn = {978-1-4051-3666-2},
	shorttitle = {Games and information},
	pagetotal = {528},
	publisher = {Blackwell Pub},
	author = {Rasmusen, Eric},
	date = {2007},
	% note = {{OCLC}: ocm71312740},
	keywords = {Game theory},
}

@inproceedings{browne_modern_2018,
	location = {Maastricht},
	title = {Modern Techniques for Ancient Games},
	isbn = {978-1-5386-4359-4},
	url = {https://ieeexplore.ieee.org/document/8490420/},
	doi = {10.1109/CIG.2018.8490420},
	eventtitle = {2018 {IEEE} Conference on Computational Intelligence and Games ({CIG})},
	booktitle = {2018 {IEEE} Conference on Computational Intelligence and Games ({CIG})},
	publisher = {{IEEE}},
	author = {Browne, Cameron},
	urldate = {2019-11-20},
	date = {2018-08},
}

@article{parlett_whats_2016,
	title = {What's a Ludeme?},
	volume = {2},
	url = {https://www.parlettgames.uk/gamester/ludemes.html},
	pages = {81--84},
	number = {2},
	journaltitle = {Game \& Puzzle Design},
	author = {Parlett, David},
	date = {2016},
}

@online{wikipedia_contributors_chopsticks_2019,
	title = {Chopsticks (hand game) — Wikipedia, The Free Encyclopedia},
	url = {https://en.wikipedia.org/w/index.php?title=Chopsticks_(hand_game)&oldid=925512510},
	author = {{Wikipedia contributors}},
	date = {2019},
    urldate = {2019-11-10}
}

@book{engelstein_building_2019,
	location = {Boca Raton, {FL}},
	title = {Building blocks of tabletop game design: an encyclopedia of mechanisms},
	isbn = {978-1-138-36549-0 978-1-138-36552-0},
	shorttitle = {Building blocks of tabletop game design},
	pagetotal = {516},
	publisher = {Taylor \& Francis},
	author = {Engelstein, Geoffrey and Shalev, Isaac},
	date = {2019}
}

@book{tekinbas_rules_2003,
	location = {Cambridge, {MA}},
	title = {Rules of play: game design fundamentals},
	isbn = {978-0-262-24045-1},
	shorttitle = {Rules of play},
	pagetotal = {672},
	publisher = {{MIT} Press},
	author = {Tekinbaş, Katie Salen and Zimmerman, Eric},
	date = {2003}
}

@book{koster_theory_2005,
	location = {Scottsdale, {AZ}},
	title = {A theory of fun for game design},
	isbn = {978-1-932111-97-2},
	pagetotal = {244},
    pages = {118},
	publisher = {Paraglyph Press},
	author = {Koster, Raph},
	date = {2005},
	% note = {{OCLC}: ocm57406861},
	keywords = {Computer games, Design, Aspect social, Conception, Jeux d'ordinateur, Social aspects},
}

@book{schell_art_2019,
	location = {Boca Raton, {FL}},
	edition = {Third edition},
	title = {The art of game design: a book of lenses},
	isbn = {978-1-138-63205-9 978-1-138-63209-7},
	shorttitle = {The art of game design},
	pagetotal = {520},
	publisher = {Taylor \& Francis},
	author = {Schell, Jesse},
	date = {2019}
}

@inproceedings{khalifa_modifying_2016,
	title = {Modifying {MCTS} for Human-like General Video Game Playing},
	isbn = {978-1-57735-770-4},
	url = {http://dl.acm.org/citation.cfm?id=3060832.3060973},
	series = {{IJCAI}'16},
	pages = {2514--2520},
	booktitle = {Proceedings of the Twenty-Fifth International Joint Conference on Artificial Intelligence},
	publisher = {{AAAI} Press},
	author = {Khalifa, Ahmed and Isaksen, Aaron and Togelius, Julian and Nealen, Andy},
	date = {2016},
	% note = {event-place: New York, New York, {USA}},
}

@article{cousins_elementary_2004,
	title = {Elementary game design},
	url = {https://benjaminjcousins.wordpress.com/2014/04/01/elementary-game-design/},
	pages = {51--54},
	journaltitle = {Develop Magazine},
	author = {Cousins, Ben},
	date = {2004-10},
}

@unpublished{koster_grammar_2005,
	location = {San Francisco, {CA}},
	title = {A Grammar of Gameplay},
	% note = {Game Developers Conference 2005},
	author = {Koster, Raph},
	date = {2005},
}

@online{stephane_game_2006,
	title = {A Game Grammar},
	url = {http://www.stephanebura.com/diagrams/},
	titleaddon = {stephanebura.com},
	author = {Stéphane, Bura},
	date = {2006-03},
    urldate = {2019-11-25},
}

@unpublished{koster_game_2015,
	title = {Game Grammar},
	url = {https://www.raphkoster.com/games/presentations/game-grammar/},
	% note = {{PAXDev}},
	author = {Koster, Raph},
	date = {2015}
}

@online{koster_atomic_2012,
	title = {An atomic theory of fun game design},
	url = {https://www.raphkoster.com/2012/01/24/an-atomic-theory-of-fun-game-design/},
	titleaddon = {Raph Koster's Website},
	author = {Koster, Raph},
	date = {2012-01-24},
    urldate = {2019-11-25},
}

@book{adams_game_2012,
	location = {Berkeley, {CA}},
	title = {Game mechanics: advanced game design},
	isbn = {978-0-321-82027-3},
	shorttitle = {Game mechanics},
	pagetotal = {353},
	publisher = {New Riders},
	author = {Adams, Ernest and Dormans, Joris},
	date = {2012},
	% note = {{OCLC}: ocn802250851},
	keywords = {Computer games, Design, Programming},
}

@online{noauthor_notitle_nodate,
	url = {https://ludii.games},
	titleaddon = {Ludii Portal},
    urldate = {2019-11-10},
}

@article{chinchalkar_upper_1996,
	title = {An Upper Bound for the Number of Reachable Positions},
	volume = {19},
	issn = {24682438, 13896911},
	url = {https://www.medra.org/servlet/aliasResolver?alias=iospress&doi=10.3233/ICG-1996-19305},
	doi = {10.3233/ICG-1996-19305},
	pages = {181--183},
	number = {3},
	journaltitle = {{ICGA} Journal},
	shortjournal = {{ICG}},
	author = {Chinchalkar, S.},
	urldate = {2019-12-04},
	date = {1996-09-01}
}

@article{rabin_probabilistic_1963,
	title = {Probabilistic automata},
	volume = {6},
	issn = {00199958},
	url = {https://linkinghub.elsevier.com/retrieve/pii/S0019995863902900},
	doi = {10.1016/S0019-9958(63)90290-0},
	pages = {230--245},
	number = {3},
	journaltitle = {Information and Control},
	shortjournal = {Information and Control},
	author = {Rabin, Michael O.},
	urldate = {2019-12-03},
	date = {1963-09},
	langid = {english},
}

@article{hammond_schumpeterian_2007,
	title = {Schumpeterian Innovation in Modelling Decisions, Games, and Economic Behaviour},
	volume = {15},
	pages = {179--195},
	number = {1},
	journaltitle = {History of Economic Ideas},
	author = {Hammond, Peter J.},
	date = {2007},
	note = {{SPECIAL} {ISSUE}: {NEW} {PERSPECTIVES} {ON} {THE} {SCHUMPETER} {FRONTIER} (2007)},
}

@book{chvatil_mage_2011,
	title = {Mage {K}night},
	publisher = {{WizKids}},
	author = {Chvátil, Vlaada},
	date = {2011},
	note = {board game}
}

@book{garfield_magic:_1999,
	title = {Magic: {T}he {G}athering {C}lassic ({S}ixth {E}dition)},
	url = {https://archive.wizards.com/magic/generic/official/Rulings/archive/rules042399.txt},
	publisher = {Wizards of the Coast},
	author = {Garfield, Richard and Rose, Bill and Moursund, Beth},
	urldate = {2019-12-06},
	date = {1999},
	note = {collectible card game}
}

@book{childres_gloomhaven_2017,
	title = {Gloomhaven},
	publisher = {Cephalofair Games},
	author = {Childres, Isaac},
	date = {2017},
	note = {board game}
}

@book{glassco_betrayal_2004,
	title = {Betrayal at House on the Hill},
	publisher = {Avalon Hill Games, Inc.},
	author = {Glassco, Bruce and Daviau, Rob and {McQuillan}, Bill and Selinker, Mike and Woodruff, Teeuwynn},
	date = {2004},
	note = {board game}
}

@book{lamorisse_risk:_1959,
	title = {Risk: {T}he {C}ontinental {G}ame},
	publisher = {Parker Brothers},
	author = {Lamorisse, Albert and Levin, Michael I.},
	date = {1959},
	note = {board game}
}

@misc{riggins_in_progress,
    author = {Riggins, Paul},
    note = {in preparation},
}

@online{rogers_6_2019,
	title = {The 6 Zones of Play},
	url = {https://mrbossdesign.blogspot.com/2019/07/the-6-zones-of-play.html},
	titleaddon = {mr. boss' design lair},
	author = {Rogers, Scott},
	urldate = {2020-01-06},
	date = {2019-07-25}
}

\appendix

\section{Agency Equivalence: Technical Definitions}
\label{sec:EquivalenceTechnicalDefinitions}

Here we will flesh out the component definitions for agency equivalence, \cref{def:AgencyEquivalence}.
There are two main parts to this: formalizing the necessary correspondences to establish game tree equivalence up to relabeling, and then formalizing the tree reductions to remove spurious differences unrelated to player agency.

\subsection{Game Tree Correspondences and Equivalence}
\label{sec:GameTreeCorrespondencesAndEquivalence}

To start, let's strip a game tree of all labels and consider what it means to match what's left:

\begin{definition}
    \label{def:StrippedGameTree}
    A \emph{stripped game tree}, denoted $\langle T \rangle$, is a game tree $T$ with all labels removed; only the arrangement of nodes and edges remains.
\end{definition}

\begin{definition}[Structural equivalence]
    \label{def:StructuralEquivalence}
    Two game trees $T,T'$ (or game systems $\mathcal{G},\mathcal{G}'$) are \emph{structurally equivalent} if the stripped trees are equal $\langle T \rangle = \langle T' \rangle$ (or if the sets of stripped game trees are equal $\langle \text{Tr}\,\mathcal{G} \rangle = \langle \text{Tr}\,\mathcal{G}' \rangle$). 

    This establishes a bijective \emph{structural correspondence} $f: n \mapsto n'$, similarly $f: e \mapsto e'$, between the labelled nodes and edges of corresponding trees $T$ and $T'$ (or $t \in \text{Tr}\,\mathcal{G}$ and $t' \in \text{Tr}\,\mathcal{G}'$).
    Several such correspondences may be possible (e.g., because a tree is symmetric).
\end{definition}

This is sufficient to say that the arrangement of nodes and edges is the same, but some of the labels do contain important content that distinguishes one game system from another in substance, not just aesthetics.
In particular, we want to see that the corresponding probabilities are the same, players have the same kinds of choices to make at each state node, and that the outcomes are similarly distinct.

Comparing probabilities and outcomes is formally straightforward:
\begin{definition}[Matching probabilities]
    \label{def:CorrespondingProbabilities}
    For each chance edge $e$ in a game tree, let $p(e)$ be the assigned probability.
    Two structurally equivalent game trees $T,T'$ (or game systems $\mathcal{G}, \mathcal{G}'$) with structural correspondence $f: T \to T'$ are said to have \emph{matching probabilities} if $p(e) = p(f(e))$ for all chance edges $e \in T$ ($\in \text{Tr}\,\mathcal{G}$).
\end{definition}

\begin{definition}[Similarly distinct outcomes]
    \label{def:CorrespondinglyDistinctOutcomes}
    Take two structurally equivalent game trees $T,T'$ (or game systems $\mathcal{G},\mathcal{G}'$) with correspondence $f$, and let $O$ and $O'$ be the sets of all distinct outcomes assigned to their respective terminal nodes.
    We say $T$ and $T'$ (or $\mathcal{G}$ and $\mathcal{G}'$) have \emph{similarly distinct outcomes} if there exists a bijective map $o: O \to O'$ (or if there exists such a map between the outcomes of every $T \in \text{Tr}\,\mathcal{G}$ and $T' = f(T) \in \text{Tr}\,\mathcal{G}'$).
\end{definition}

Note that the set of distinct outcomes $O$ assigned to terminal nodes may be different than the set of all possible outcomes $\mathcal{O}$, if a game system contains possible outcomes that are never reached in a particular tree.

Confirming that players have the same kinds of decisions along the way is more involved, at least formally.
We want single-player nodes to still be single-player nodes with the same number of choices, and multiplayer nodes to still be multiplayer nodes with the same interaction between each player's simultaneous decisions.
In essence, we want the same strategic form game to be played at each internal state node, as described in \cref{sec:GameTreesAndAutomata}.
First let us define the decision matrix, which is the strategic form game at each node:
\begin{definition}[Decision matrix]
    \label{def:DecisionMatrix}
    Let $w$ be a non-terminal state node in a game tree with assigned state $s$ and outgoing decision edges $E$.
    There is a nonempty set of \emph{active players} $\{ p_1, \ldots, p_k \} \subseteq \mathcal{P}$ that can make legal decisions at this state, i.e., they have nonempty legal sets $L_{p_i}(s)$.
    The \emph{decision matrix} at node $w$ is a map $D_w: L_{p_1}(s)\times\cdots\times L_{p_k}(s)\to E$ of active decision tuples to edges.

Game tree reductions or transformations may adjust the domain of $D_w$ (e.g., see \cref{def:SinglePlayerSubtree,def:DecisionMatrixRedundancy}).
More generally, each active player $p$ has a set of legal choices $\ell_p(w)$ they can select to influence the edge followed, such that $D_w: \ell_{p_1}(w)\times\cdots\times\ell_{p_k}(w) \to E$.
A game tree produced freshly from \cref{def:GameSystemTree} has $\ell_p(w) = L_p(s)$.
\end{definition}

Some sample decision matrices are illustrated in \cref{fig:DecisionMatrixIllustration,fig:MatchingDecisionTreesExample,fig:DecisionMatrixIllustrationRelabeled}.

In a game tree or reduced game tree where each edge $e \in E$ is labeled with a set of one or more unique decision tuples, $D_w$ simply maps each active decision tuple $d^k = (d_{p_1}, \ldots, d_{p_k})$ to the edge with its corresponding decision tuple $d_0^n$ (in which each active player $p_i$ chooses $d_{p_i}$, and non-active players are assigned the null decision.)
An edge can obtain multiple decision tuples, even though the game tree construction \cref{def:GameSystemTree} only assigns one to each edge, due to something like a symmetry-redundant subtree reduction (see \cref{def:SymmetryRedundantSubtree}).

To illustrate, here is a sample decision matrix with outgoing edges as it might appear in a reduced game tree, for a 3-player game with $\mathcal{P} = (\text{P1},\text{P2},\text{P3})$. 
Players P1 and P3 are the active players:
\def\gameX{
    \mbox{
    \setlength\tabcolsep{1.5pt}
    \begin{tabular}{cc|c|c|c|cc}
                            & \multicolumn{1}{c}{} & \multicolumn{3}{c}{P3}    &                           &                           \\
                            & \multicolumn{1}{c}{} & \multicolumn{1}{c}{$d_2$} & \multicolumn{1}{c}{$d_3$} & \multicolumn{1}{c}{$d_4$} &                  &              \\ \cline{3-5}
        \multirow{2}*{P1}   & $d_1$                &                           &                           &                           &  \phantom{$d_1$} &              \\ \cline{3-5}
                            & $d_2$                &                           &                           &                           &  \phantom{$d_2$} & \phantom{P1} \\ \cline{3-5}
    \end{tabular}
}}\def\gameY{
    \mbox{
    \setlength\tabcolsep{3pt}
    \begin{tabular}{cc|c|c|c|cc}
                             & \multicolumn{1}{c}{} & \multicolumn{3}{c}{P3}  &                         &                         \\
                             & \multicolumn{1}{c}{} & \multicolumn{1}{c}{$c$} & \multicolumn{1}{c}{$d$} & \multicolumn{1}{c}{$e$} &                  &              \\ \cline{3-5}
        \multirow{2}*{P1}    & $a$                  & $\alpha$                & $\gamma$                & $\alpha$                &  \phantom{$d_1$} &              \\ \cline{3-5}
                             & $b$                  & $\gamma$                & $\beta$                 & $\gamma$                &  \phantom{$d_2$} & \phantom{P1} \\ \cline{3-5}
    \end{tabular}
}}
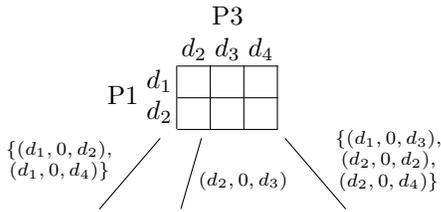
\begin{figure}[H]
    \centering
    \vspace{-8pt}
    \begin{forest} 
        for children={tier=a}
        [\gameX
            [\phantom{$\{ \}$},font=\scriptsize,
                edge label={node[midway,left,font=\scriptsize,
                    xshift=-3pt,yshift=5pt,
                    align=center]
                    {$\{ (d_1,0,d_2),$\\ $(d_1,0,d_4) \}$}}]
            [\phantom{$(d_2,0,d_3)$},l=1.5cm,font=\scriptsize,
                edge label={node[pos=0.6,right,font=\scriptsize]
                {$(d_2,0,d_3)$}}]
            [\phantom{$\{ (d_1,0,d_2), (d_1,0,d_4) \}$},font=\scriptsize,
                edge label={node[midway,right,font=\scriptsize,
                    xshift=4pt,yshift=5pt,
                    align=center]
                    {$\{ (d_1,0,d_3),$\\ $(d_2,0,d_2),$\\ $(d_2,0,d_4) \}$}}]
        ]
    \end{forest}
    \vspace{-15pt}
    \caption{Sample illustration of a decision matrix.}
    \vspace{-5pt}
    \label{fig:DecisionMatrixIllustration}
\end{figure}

An alternative labeling might make the structure of the joint decisions more clear:
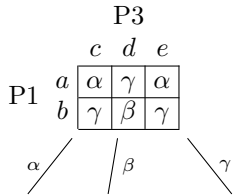
\begin{figure}[H]
    \centering
    \vspace{-8pt}
    \hspace{0pt}
    \begin{forest} 
        for children={tier=a}
        [\gameY
            [\phantom{$\{ \}$},font=\scriptsize,
                edge label={node[midway,left,font=\scriptsize,
                    align=center]
                    {$\alpha$}}]
            [\phantom{$(d_2,0,d_3)$},l=1.3cm,font=\scriptsize,
                edge label={node[pos=0.5,right,font=\scriptsize]
                {$\beta$}}]
            [\phantom{$\{ (d_1,0,d_2),\}$},font=\scriptsize,
                edge label={node[midway,right,font=\scriptsize,
                    align=center]
                    {$\gamma$}}]
        ]
    \end{forest}
    \vspace{-15pt}
    \caption{Relabeling of the decision matrix in \cref{fig:DecisionMatrixIllustration}.}
    \vspace{-5pt}
    \label{fig:DecisionMatrixIllustrationRelabeled}
\end{figure}

Colloquially, we want to say that two decision matrices match (in the context of a game tree) if there is a self-consistent way to relabel the players, decisions, and edges such that the relabeled players making the relabeled decisions lead to the relabeled edges.
Formally:
\begin{definition}[Matching decision matrices]
    \label{def:MatchingDecisionMatrices}
    Take two game trees $T$ and $T'$ that are structurally equivalent with structural correspondence $f: T \to T'$.
    Let $w \in T$ and $w' = f(w) \in T'$ be corresponding non-terminal state nodes, with states $s, s'$, decision matrices $D_w,D_{w'}$, and sets of active players $P, P'$, respectively.

    If possible, define a bijective \emph{active player correspondence} $g: P \to P'$ between the sets of active players, with associated bijective maps $h_{g,p}: \ell_p(w) \to \ell_{g(p)}(w')$, $p \in P$, between each pair of legal choices that correspond in the different games.
    Taken together, these furnish a unique map between the active decision tuples $h_g: d^k \mapsto d'^k$, i.e., between the domains of $D_w$ and $D_{w'}$.

    The decision matrices $D_w$ and $D_{w'}$ are said to \emph{match}, denoted $D_w \sim D_{w'}$, if there exists at least one such $g$ and set $\{h_{g,p}\}$ such that the matrices map to corresponding edges: i.e., $D_w(d^k) = e \in T$ and $D_{w'}(h_g(d^k)) = f(e) \in T'$ for all $d^k$ in the domain of $D_w$.
\end{definition}
See \cref{fig:MatchingDecisionTreesExample} for examples.
\cref{fig:DecisionMatrixIllustration,fig:DecisionMatrixIllustrationRelabeled} could also be said to match, since they only differ in choice labeling, if the node and edges were placed to correspond in two structurally equivalent trees.
Note that we may relabel the decisions for each player separately.
If two active players choose the same decision in one matrix (e.g., $(d_2,d_2)$ in \cref{fig:DecisionMatrixIllustration}), but different decisions in a matching matrix (e.g., $(b,c)$ in \cref{fig:DecisionMatrixIllustrationRelabeled}), we do not care as long as the resulting edges correspond.
Those are unimportant differences in labeling.

\captionsetup[figure]{skip=0pt}
\newcommand\bigredsim{\stackrel{\mathclap{\normalfont\mbox{\footnotesize agency}}}{\scalebox{3.5}{$\sim$}}}
\newcommand\redsim{\mathrel{\overset{\makebox[0pt]{\mbox{\normalfont\tiny red.}}}{\sim}}}
\begin{figure*}[!tb]
    \centering
    \setlength\tabcolsep{8pt}
    \begin{tabular}{m{.38\linewidth} m{.04\linewidth} m{0.45\linewidth}}
        \hspace{-23pt}
    \includegraphics[scale=0.95]{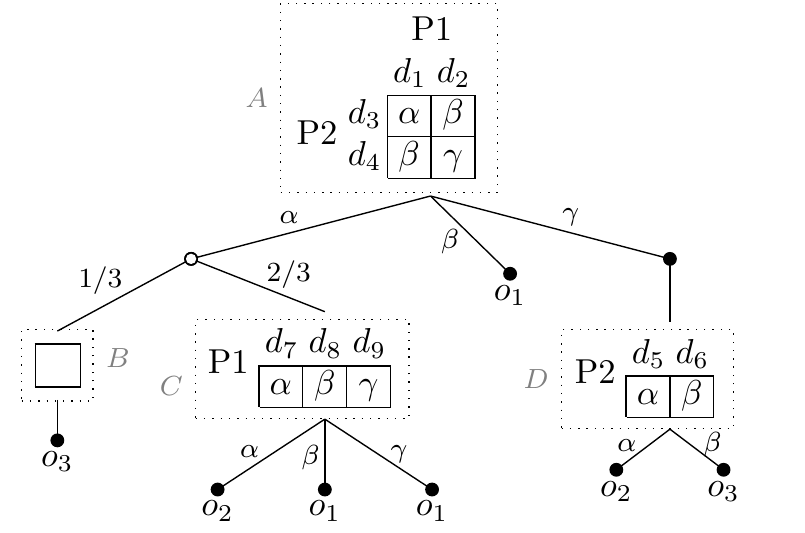}
    &
    $\bigredsim$
    &
        % \hspace{-23pt}
    \includegraphics[scale=0.95]{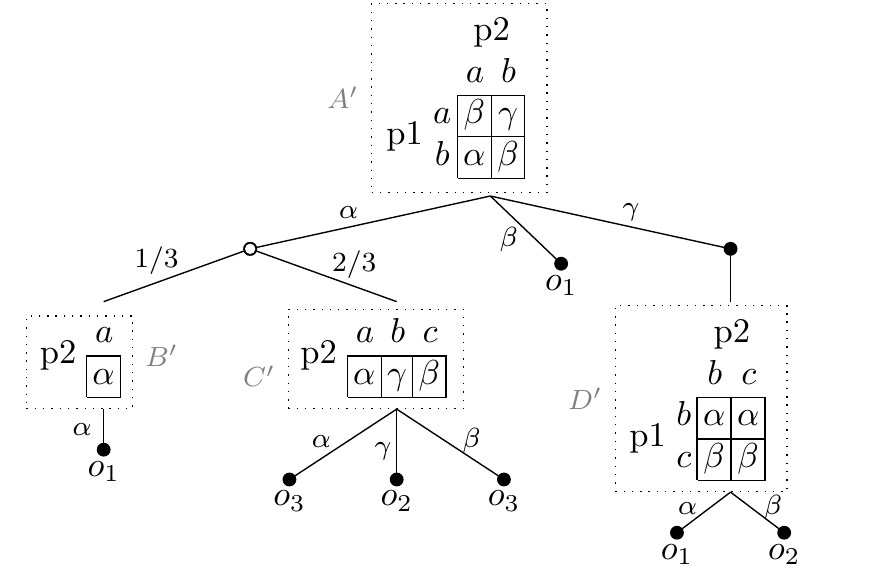}
    \end{tabular}
    \caption{
        These are two full game trees, with state labels suppressed, outcomes $o_i$ on each terminal node, probabilities ($1/3$ and $2/3$) on chance edges, and decision edges labeled with decision matrix outcomes instead of sets of decision tuples, for brevity (e.g., as \cref{fig:DecisionMatrixIllustrationRelabeled} relabels \cref{fig:DecisionMatrixIllustration}).
        All decision matrices are illustrated on non-terminal state nodes.
        Some (but not all) of the decision matrices match: $A \sim A'$, $B \not\sim B'$, $C \sim C'$, and $D \not\sim D'$.
        However, all of them match after reduction by \cref{def:DecisionMatrixRedundancy}: in particular, $B \redsim B'$, $D \redsim D'$ (the blank matrix $B$ is a decision matrix with empty domain).
        In fact, $B$ ($D$) is the reduced form of $B'$ ($D'$), after some relabeling.
        Thus, after these reductions, the \emph{trees have matching decision matrices} (\cref{def:TreesWithMatchingDecisionMatrices}) under the player correspondence P1 $\leftrightarrow$ p2 and P2 $\leftrightarrow$ p1.
        Incidentally, because the probabilities also correspond and the outcomes are similarly distinct, these trees are equivalent up to relabeling (\cref{def:GameTreeEquivalenceUpToRelabeling}) after appropriate reductions: i.e., the two trees are agency equivalent (\cref{def:AgencyEquivalence}).
    }
    \label{fig:MatchingDecisionTreesExample}
\end{figure*}

Note also that if an overall \emph{player correspondence} $\pi: \mathcal{P} \to \mathcal{P}'$ is given between all the players of two trees, this fixes a unique active player correspondence $g$ at any given node.
We use this to generalize beyond individual decision matrices to game trees and systems:
\begin{definition} \textbf{(Trees with matching decision matrices)}
    \label{def:TreesWithMatchingDecisionMatrices}
    Take two structurally equivalent game trees $T,T'$ (or game systems $\mathcal{G},\mathcal{G}'$) with structural correspondence $f: T \to T'$ and sets of players $\mathcal{P}$ and $\mathcal{P}'$. 
    We say these trees (or systems) have \emph{matching decision matrices} if there exists at least one bijective player correspondence $\pi: \mathcal{P} \to \mathcal{P}'$ such that all corresponding decision matrices match with respect to $\pi$---%
    i.e., $D_w \sim D_{f(w)}$ for all internal state nodes $w \in T\ ( \in \mathcal{G})$ with each active player correspondence $g$ fixed by $\pi$. (See \cref{def:MatchingDecisionMatrices}.
    The associated decision mappings $\{h_{g,p}\}$ need not be compatible across all nodes in the tree.)
\end{definition}
That is, whatever relabeling we have to do to see that the decision matrices match, we want at least the player relabeling to be consistent across all the decision matrices.
This is illustrated in \cref{fig:MatchingDecisionTreesExample}.
We do not demand the same for the decision relabelings: the decision $d$ may be relabeled (by $h_{g,p}$) to $d'_1$ for player $p$ at one node, but to $d'_2\neq d'_1$ at another, and that is fine as long as \cref{def:MatchingDecisionMatrices} is satisfied at each.

Finally, let us put all of this together to give a broadly useful sense of equivalence between game trees:
\begin{definition}[Equivalence up to relabeling]
    \label{def:GameTreeEquivalenceUpToRelabeling}
    We say that two game trees $T,T'$ are \emph{equivalent up to relabeling} (or that game systems $\mathcal{G},\mathcal{G}'$ are \emph{game tree equivalent up to relabeling}) if $T$ and $T'$ ($\mathcal{G}$ and $\mathcal{G}'$) are structurally equivalent and have matching probabilities, matching decision matrices, and similarly distinct outcomes, all with respect to the same structural correspondence $f$.
\end{definition}
This sense of equivalence for game trees or systems respects everything about them except for the specific labels chosen to represent players, states, decisions, and outcomes.

If some of those labels are additionally identical, we might say (using outcomes as an example) that two trees $t$ and $t'$ are equivalent up to relabeling and with identical outcomes.
This means that if $t$ and $t'$ have structural correspondence $f: t \to t'$ and $\omega(z)$ gives the outcome assignment of terminal node $z \in t$, then $\omega(z) = \omega(f(z))$ for all $z$.

If all labels additionally happen to be identical, then we might simply call the game trees \emph{equivalent}, or the game systems \emph{game tree equivalent}.

It is worth noting that \cref{def:GameTreeEquivalenceUpToRelabeling}, in not distinguishing between the content of outcome labels, does not distinguish between whether an outcome might be good or bad for a player---we have chosen to leave such value judgments up to the player models (see \cref{sec:GameDescriptions}).
The normal and \emph{mis\`ere} versions of a game have opposite win/lose conditions, for instance, but would be considered equivalent up to relabeling.

\subsection{Game Tree Reductions}
\label{sec:GameTreeReductions}

To establish agency equivalence from \cref{def:AgencyEquivalence}, we need to prune those differences between trees that are not meaningful from the standpoint of player agency.
Here we describe how to perform the relevant transformations to reduce bookkeeping subtrees, single-player subtrees, symmetry-redundant subtrees, and decision matrix redundancies, as heuristically described in \cref{sec:AgencyEquivalence}.

\captionsetup[figure]{skip=-5pt}
\begin{figure*}[!tb]
    \centering
    \begin{tabular}{m{.44\linewidth} m{.06\linewidth} m{0.45\linewidth}}
    \includegraphics[scale=0.95]{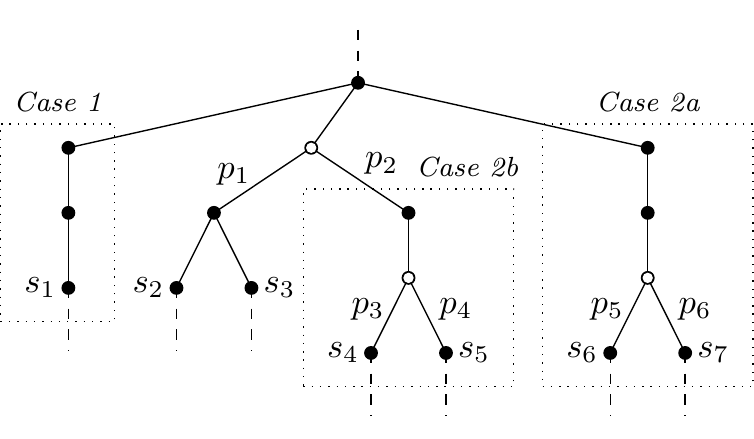}
    &
    \tikz \draw[-{latex},line width=1mm] (0,0) -- (1,0);
    &
    \includegraphics[scale=0.95]{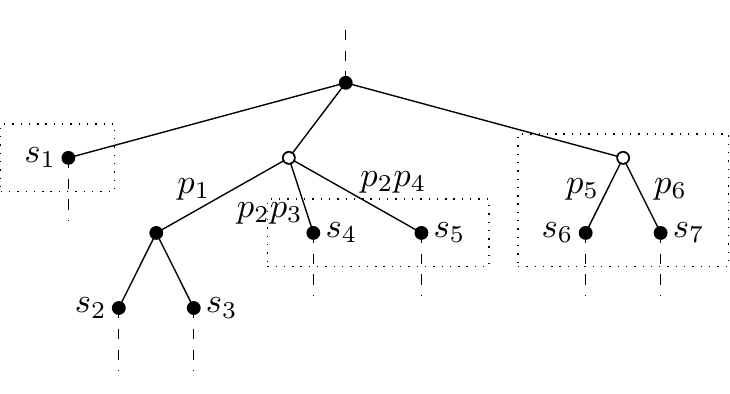}
    \end{tabular}
    \caption{
        Bookkeeping subtree reduction example. 
        Dotted lines highlight the bookkeeping subtrees before and after being reduced, exemplifying the different cases in \cref{def:BookkeepingSubtree}.
        Solid dots are state nodes, circles are chance nodes, and chance edges are labeled with probabilities $p_i$.
        The labels on most state nodes and all decision edges have been omitted.
        Dashed lines connect to other parts of the game tree.
    }
    \label{fig:BookkeepingReductionExample}
\end{figure*}

The following notion will be useful for these definitions:
\begin{definition}
    \label{def:GameTrajectory}
    A \emph{(partial) legal game trajectory} is a path along directed edges between any two nodes in a game tree.
    It is called \emph{full} instead of partial if it extends from the root node to a terminal node.
\end{definition}

Each full legal game trajectory captures a playthrough that may be generated by the gameplay algorithm \cref{def:GameplayAlgorithm}.

\begin{figure*}[tb!]

    \centering
    \hspace{-40pt}
    \begin{tabular}{m{.42\linewidth} m{.01\linewidth} m{0.45\linewidth}}
    \includegraphics[scale=0.90]{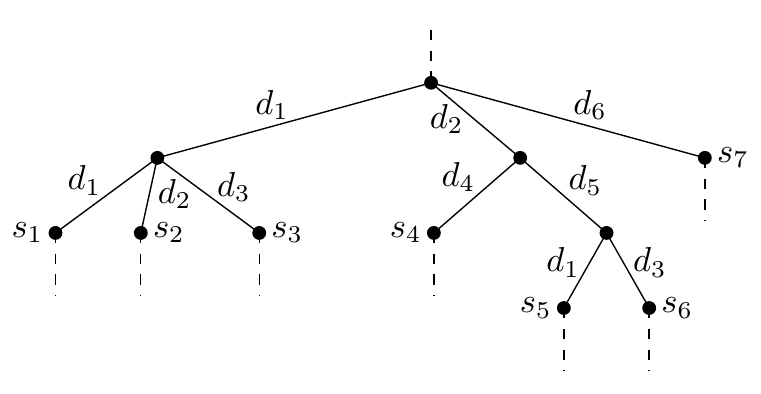}
    &
    \tikz \draw[-{latex},line width=1mm] (0,0) -- (1,0);
    &
    \includegraphics[scale=0.90]{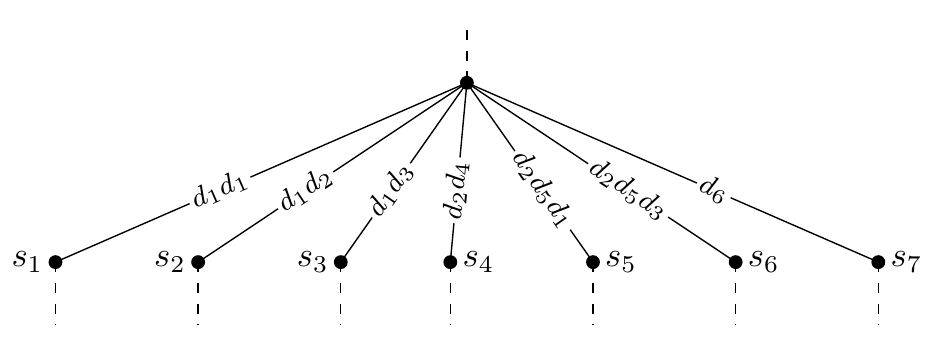}
    \end{tabular}
    \caption{
        Single-player subtree reduction example, illustrating \cref{def:SinglePlayerSubtree}.
        All state nodes, except possibly the leaves, belong to the same player.
        Internal state node labels have been omitted.
        Edges have been labeled only with this player's decisions, for brevity, since all other players take the null decision everywhere.
        Dashed lines connect to other parts of the game tree.
    }
    \label{fig:SinglePlayerReductionExample}
\end{figure*}

Now, on to the reductions, which are also illustrated in \cref{fig:MatchingDecisionTreesExample,fig:BookkeepingReductionExample,fig:SinglePlayerReductionExample}:

\begin{definition}
    \label{def:BookkeepingSubtree}
    A \emph{bookkeeping subtree} is a subtree of a game tree rooted at a state node $r$ and with state nodes as leaves, which has exactly one decision edge proceeding out of $r$ and each interior state node.
    Players are unable to influence play in this subtree.
    A bookkeeping subtree may be reduced as follows:\medskip

    \emph{Case 1:} If there are no chance nodes in the subtree:
        \begin{enumerate}
            \item There is only a single leaf, with state $s$.
                Replace the entire subtree with a state node with state $s$.
        \end{enumerate}

    \emph{Case 2:} If there are chance nodes in the subtree:
    \begin{enumerate}
        \item Let $G$ be the set of all partial game trajectories from $r$ to the subtree leaves.
            Let $l(g)$ be the final state node in each $g\in G$ (i.e., the leaves).
        \item Assign each trajectory $g$ a probability $p(g)$ given by the product of the probabilities on the chance edges in $g$.
        \item Then, if the parent $r'$ of the subtree root $r$ is \ldots
            \begin{enumerate}
                \item \emph{Case 2a:} \ldots a state node: 
                    Replace $r$ with a new chance node $c$.
                \item \emph{Case 2b:} \ldots a chance node $c$\,: proceed to step \ref{step:DeleteAllNodes}. 
                    (Note $r$ has an incoming chance edge with probability $p_r$.)
                \item \emph{Case 2c:} \ldots nonexistent ($r$ is the root of the game tree): 
                    Replace the child of $r$ with a new chance node $c$.
            \end{enumerate}
        \item \label{step:DeleteAllNodes} Delete all nodes and edges between $c$ and the leaves, non-inclusive.
        \item Draw new chance edges between $c$ and each leaf $l(g)$, labeled by the corresponding probabilities $p(g)$, or $p_r\cdot p(g)$ in Case 2b.
    \end{enumerate}
\end{definition}
Bookkeeping reductions are illustrated in \cref{fig:BookkeepingReductionExample}.

\captionsetup[figure]{skip=-9pt}

\begin{definition}
    \label{def:SinglePlayerSubtree}
    A \emph{single-player deterministic subtree} is a subtree of a game tree rooted at a state node $r$ and with state nodes as leaves, without any chance nodes, in which all nodes belong to a single player (except perhaps the leaves).
    Only that player has any meaningful decisions in this subtree, and they could just as well be made all at once.
    Such a subtree may be reduced as follows:
    \begin{enumerate}
        \item Let $G$ be the set of all partial game trajectories from $r$ to its leaves.
            Let $l(g)$ be the final state node in each $g\in G$ (the leaves).
        \item Delete all nodes and edges between $r$ and the leaves, non-inclusive.
        \item Draw a new decision edge between $r$ and each leaf $l(g)$, labeled by the sequence of decision tuples in $g$.
    \end{enumerate}
    A single-player subtree reduction is illustrated in \cref{fig:SinglePlayerReductionExample}.

    This reduction also changes the domain of the decision matrix $D_r$ at $r$ (see \cref{def:DecisionMatrix}):
    The legal choices $\ell_p(r)$ available to the single active player $p$ at $r$ are now taken to be the set of decision tuple sequences for each $g \in G$ (i.e., the set of labels on the new decision edges proceeding from $r$), \emph{not} the canonical legal set $L_p$.
\end{definition}

\begin{definition}
    \label{def:SymmetryRedundantSubtree}
    A \emph{symmetry-redundant subtree} is a subtree $t$ of a game tree rooted at a state or chance node $r$ and extending to all its descendants, that is equivalent up to relabeling, and with identical outcomes, to a subtree $t'$ rooted at a sibling node $r'$ (and extending to all of its descendants).

    A lone terminal node $z$ is also considered a symmetry redundant subtree if its outcome $\omega(z)$ is identical to one of its siblings.

    The symmetry redundant subtree $t$ may be reduced as follows:
    \begin{enumerate}
        \item Let $e,e'$ be the edges ingoing to $r,r'$, respectively.
            If $e$ and $e'$ are \ldots
            \begin{enumerate}
                \item \emph{Case 1:} \ldots decision edges with assigned tuples sets $d(e)$ and $d(e')$ (if only one tuple, consider it a set of size 1).
                    Replace the tuple set on $e'$ with $d(e')\cup d(e)$.
                \item \emph{Case 2:} \ldots chance edges with assigned probabilities $p(e)$ and $p(e')$:
                    Replace the probability on $e'$ with $p(e') + p(e)$.
            \end{enumerate}
        \item Delete the entire subtree $t$ rooted at $r$, and the edge $e$.
        \item (For \emph{Case 2:}) If $e'$ now has assigned probability 1, this chance edge is superfluous, as is the chance node parent $c'$ of $r'$. 
            Delete $e'$, and replace $c'$ by moving $r'$ (with subtree attached) into its place.
    \end{enumerate}
\end{definition}
Note symmetry-redundant subtrees may commonly occur when the sibling nodes $r$ and $r'$ are assigned the same state, e.g., when two different decisions from the parent state lead to the exact same game state.

Several decision tuples may end up on a single edge (e.g., \cref{fig:DecisionMatrixIllustration}) because of symmetry-redundant subtree reductions, which may lead to redundancies in decision matrices---if they were not redundant already.
To establish agency equivalence, we wish to eliminate meaningless redundancies in decision matrices.
Consider \cref{fig:DecisionMatrixIllustrationRelabeled}, for instance, which relabels \cref{fig:DecisionMatrixIllustration}. 
From this it is clear that P3 gains no additional agency by having choice $e$ available in addition to choice $c$.
Similarly, a decision matrix in a bookkeeping subtree (a matrix with only one outgoing edge) would be trivial for all players; it should not matter which player(s) are given the task of executing the bookkeeping.

We can reduce these decision matrices by removing duplicate rows and columns:
\begin{definition}
    \label{def:DecisionMatrixRedundancy}    
    A \emph{decision matrix redundancy} occurs when a decision matrix $D_w: \ell_{p_1}\times\cdots\times\ell_{p_k} \to E$ contains more than one choice for some player $p$ which lead to the same result.
    That is, when there exist distinct choices $a,b \in \ell_{p_i}$ for some $p_i$ such that  $D_w(d_1,\ldots,d_{i-1},a,d_{i+1},\ldots,d_k) = D_w(d_1,\ldots,d_{i-1},b,d_{i+1},\ldots,d_k)$ for all possible choices $d_j \in \ell_{p_j}$.
    The choices $a$ and $b$ are redundant.

    A \emph{reduced decision matrix} can be produced as follows to eliminate redundancies and unnecessary bookkeeping distinctions:
    \begin{enumerate}
        \item If there exist two redundant choices $a,b \in \ell_{p_i}$ for some $p_i$, delete one: $\ell_{p_i} \to \ell_{p_i} \setminus \{ b \}$.
        Repeat until no redundancies remain for any player.
        \item If any player $p$ has only one choice remaining ($|\ell_{p}|=1$), remove them from the active players: $P \to P \setminus p$.
        \item If no active players remain ($P = \varnothing$), there must only be a single edge $e$ in the image of $D_w$.
        Define $D_w: \varnothing \to \{ e \}$.
    \end{enumerate}
    Any two corresponding decision matrices with empty domains are said to match in the sense of \cref{def:MatchingDecisionMatrices}.

    An example of reduced matrix matching is illustrated in \cref{fig:MatchingDecisionTreesExample}.
\end{definition}

\end{document}